\documentclass{article}
\usepackage{amssymb}

\usepackage{graphicx}
\usepackage{amsmath}
\usepackage[ruled,vlined]{algorithm2e}
\usepackage[most]{tcolorbox}
\usepackage{booktabs}
\usepackage{multirow}
\usepackage{wrapfig}

\renewcommand{\O}{\mathcal{O}}

\renewcommand{\S}{\mathcal{S}} 
\newcommand{\A}{\mathcal{A}}
\newcommand{\T}{\mathcal{T}}

\newcommand{\C}{\mathcal{C}}

\newcommand{\D}{\mathcal{D}}

\newcommand{\type}{\lambda}

\newcommand{\operator}{\omega}

\newcommand{\ground}{\underline}
\newcommand{\sampler}{\sigma}

\newcommand{\oparguments}{\overline{v}}
\newcommand{\preconditions}{P}
\newcommand{\addeffects}{E^+}
\newcommand{\deleteeffects}{E^-}

\newenvironment{tightlist}%
{\begin{list}{$\bullet$}{%
    \setlength{\topsep}{0in}
    \setlength{\partopsep}{0in}
    \setlength{\itemsep}{0in}
    \setlength{\parsep}{0in}
    \setlength{\leftmargin}{1.5em}
    \setlength{\rightmargin}{0in}
}
}%
{\end{list}
}

\usepackage[preprint]{corl_2025} 

\title{From Pixels to Predicates: Learning Symbolic World Models via Pretrained Vision-Language Models}

\author{
\textbf{Ashay Athalye}\thanks{Equal contribution. Correspondence to \{ashay, njk\}@mit.edu.}\hspace{0.4em}$^{,1,4}$\;,
\textbf{Nishanth Kumar\footnotemark[1]}\hspace{0.4em}$^{,1}$, 
\textbf{Tom Silver$^2$,}
\textbf{Yichao Liang$^3$,}\\
\textbf{Jiuguang Wang$^4$,}
\textbf{Tomás Lozano-Pérez$^1$,} 
\textbf{Leslie Pack Kaelbling$^1$} 
\vspace{7px}
\\ 
$^1$MIT, $^2$Princeton University, $^3$University of Cambridge, $^4$RAI Institute \\
}

\begin{document}
\maketitle


\DeclareRobustCommand{\O}{\ensuremath{\mathcal{O}}}

\newcommand{\ours}{\texttt{pix2pred}}
\newcommand{\added}[1]{#1}


\begin{abstract}
Our aim is to learn to solve long-horizon decision-making problems in complex robotics domains given low-level skills and a handful of short-horizon demonstrations containing sequences of images.
To this end, we focus on learning abstract symbolic world models that facilitate zero-shot generalization to novel goals via planning. 
A critical component of such models is the set of symbolic \textit{predicates} that define properties of and relationships between objects.
In this work, we leverage pretrained vision-language models (VLMs) to propose a large set of visual predicates potentially relevant for decision-making, and to evaluate those predicates directly from camera images.
At training time, we pass the proposed predicates and demonstrations into an optimization-based model-learning algorithm to obtain an abstract symbolic world model that is defined in terms of a compact subset of the proposed predicates. 
At test time, given a novel goal in a novel setting, we use the VLM to construct a symbolic description of the current world state, and then use a search-based planning algorithm to find a sequence of low-level skills that achieves the goal. 
We demonstrate empirically across experiments in both simulation and the real world that our method can generalize aggressively, applying its learned world model to solve problems with a wide variety of object types, arrangements, numbers of objects, and visual backgrounds, as well as novel goals and much longer horizons than those seen at training time. 
\end{abstract}

\keywords{Planning, Learning from Demonstration, Vision-Language Models} 

\section{Introduction}
\label{sec:intro}
A core research objective in robotics is to make long-term decisions from low-level sensory inputs to accomplish a very broad range of tasks~\citep{pix2predbook}.
For instance, a useful household robot should be able to satisfy a variety of requests, such as ``wipe and clear my dining table'', ``make me a cup of healthy juice'', or even ``cook me a protein-rich meal'', across a wide range of novel household environments.
Even when provided with a library of chainable short-horizon parameterized skills (such as \texttt{moveto}, \texttt{pick}, \texttt{place}, \texttt{wipe}, \texttt{turn-knob}), solving such tasks remains challenging. The robot must not only select and sequence the right skills, but also ground each one with the correct parameters—identifying which objects to act upon, where to move, how to grasp, where to place, or how far to turn a knob.
Critically, the space of possible variations in objects, configurations, visual backgrounds, and goal specifications is extremely large: even seemingly simple instructions can require substantially different skill sequences and parameters in different situations.

In this work, we choose to tackle these challenges in a learning-from-demonstration setting with an aggressively model-based approach.
Given a handful of demonstrations that use known parameterized skills, we learn \emph{symbolic world models}~\citep{silver2023predicate,liang2024visualpredicatorlearningabstractworld} composed of abstract, interpretable properties and object relations (i.e., \textit{predicates}) that ground directly into low-level sensory inputs (i.e., \emph{pixels}). Because these models capture task-agnostic world dynamics, they can generalize across objects, embodiments (e.g., from human to robot), and tasks (e.g., from simple settings at training time to more complex, novel settings at test time).
Additionally, the structure of the models affords efficient search-based task planning~\citep{McDermott1998PDDLthePD,fox2003pddl2} to solve novel tasks at inference time~\citep{helmert2006fast}.

\begin{figure*}[!t]  
\centering
\includegraphics[width=\textwidth,trim={0 0 0 0},clip]{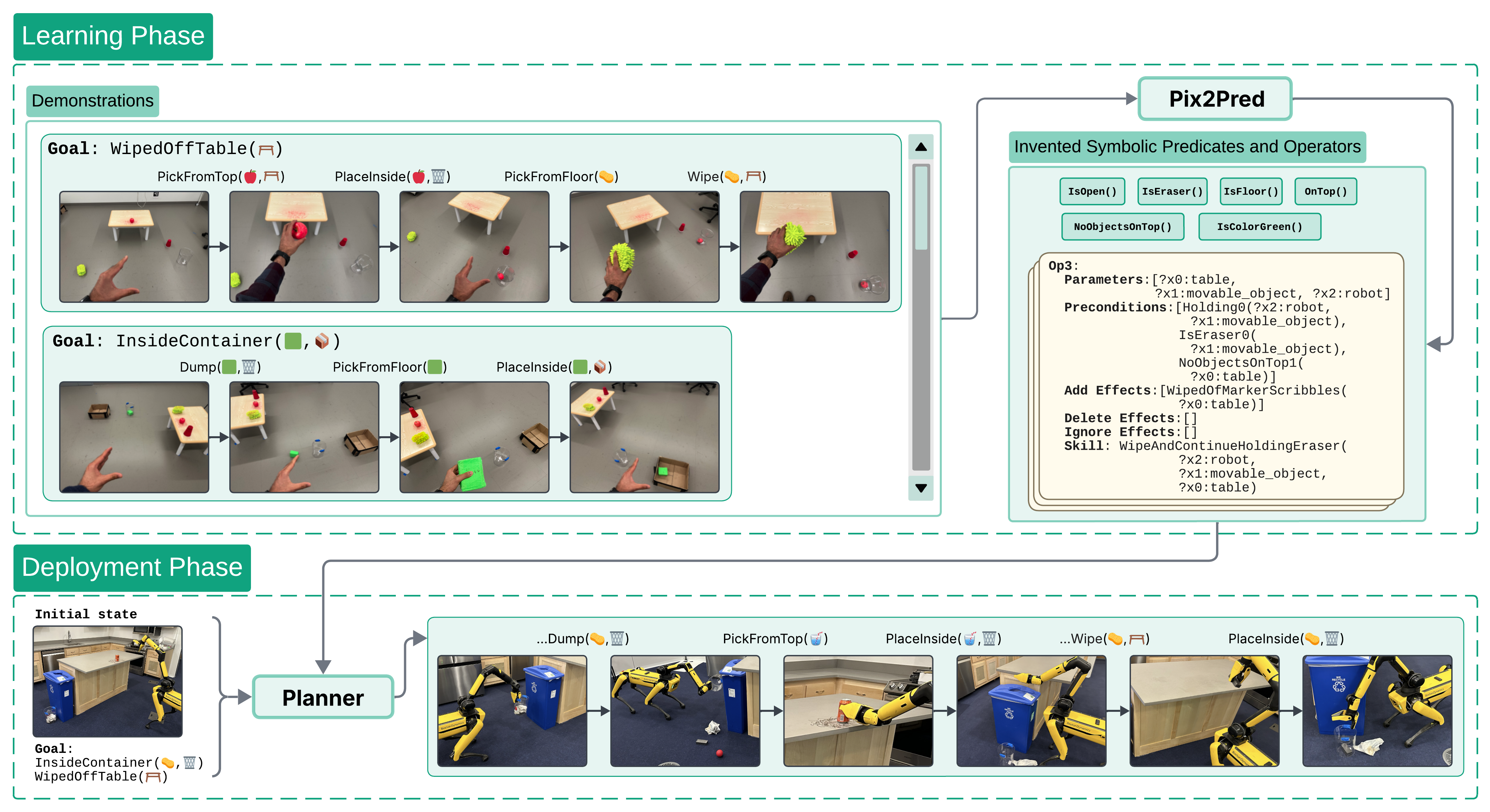}
\caption{\textbf{Overview of \ours{} in the Cleanup domain}: Given $6$ human demonstrations showcasing the effects of distinct skills (e.g., wiping, dumping) and a small initial predicate set, \ours{} invents new predicates (e.g., \texttt{NoObjectsOnTop(?table)}) and learns symbolic operators. At test time, it uses a search-based planner that operates over the learned model to solve a novel multi-step task in a visually distinct environment—e.g., retrieving an eraser from a bin, clearing an obstacle, wiping the table, and returning the eraser.}
\vspace{-3mm}
\label{fig:cleanup-domain-overview}
\end{figure*}

As an illustrative setting, consider the Cleanup domain shown in Figure~\ref{fig:cleanup-domain-overview}. At training time, the robot receives a few egocentric visual demonstrations of a human sequencing available skills --- such as grasping, dumping, and wiping --- to accomplish two specific goals: wiping a table, and throwing away a particular object. At inference time, it must achieve a novel goal of wiping, \textit{and} putting away the eraser afterward in an entirely different room where the eraser is initially in a bin (and thus the robot must first dump the bin before using the eraser and then putting it back in the bin).
This sequencing is challenging for model-free imitation learning as it requires producing an entirely different action sequence in a scenario with novel objects, and backgrounds.
However, if the agent can learn and reason over a small set of conceptual distinctions in the world state, including whether an object is an eraser or a surface is clear, then it can compose these via planning to solve unseen tasks.

Learning predicates for symbolic world models~\citep{konidaris2018skills,silver2023predicate,han2024interpretinteractivepredicatelearning,liang2024visualpredicatorlearningabstractworld} in this setting is challenging for two reasons.
\added{First, we are provided no initial information or explicit constraints on the predicates to be learned:} we must decide the number of predicates, the types and numbers of objects they should operate over, as well as their implementation (i.e, how they process input visual information).
Second, we must invent predicates that lead to a symbolic world model that generalizes well.
There are an infinite number of possible world models that could describe the dynamics illustrated by a set of demonstrations: we wish to select one that enables efficient planning across the entire distribution of tasks the demonstration tasks are drawn from.

To address these challenges, we introduce a new predicate-invention method --- \ours{} --- that exploits the perceptual grounding and common-sense capabilities of Vision-Language Models (VLMs) within a program synthesis framework.
Our method \added{assumes as input} demonstrations annotated by \added{intuitively-named} parameterized skills \added{and objects} and a small set of initial predicates primarily used to describe the demonstrations' goals.
From these, it generates a diverse pool of candidate predicates --- both programmatic, feature-based predicates (as in~\citep{silver2023predicate}) and VLM-derived, pixel-grounded predicates that capture high-level, semantically meaningful concepts (e.g., \texttt{NoObjectsOnTop(?table)}).
Given this pool, we run a subselection procedure that learns symbolic operators and selects the predicate and operator set that optimizes an efficient planning objective~\citep{silver2023predicate}.
This generate-then-subselect procedure allows \ours{} to filter redundant or unreliable predicates and retain only those that accurately ground in pixels and are useful for downstream decision-making.

In experiments, we find that \ours{} is able to learn symbolic world models that operate over pixels from just a handful ($<15$) of egocentric human demonstrations (Figure~\ref{fig:cleanup-domain-overview}).
These models enable a real-world Boston Dynamics Spot robot to solve two complex, long-horizon tasks with different layouts, objects, and visual appearances.
We leverage three simulated domains developed in different previous works to rigorously benchmark our approach's performance against several ablations and baselines from the literature.
From this, we find that \ours{} achieves the highest success rates on novel problem instances involving more objects, more complex goals, and longer horizons.

\section{Related Work}
\label{sec:related_work}
\textbf{Decision-Making with Foundation Models.} Our work is inspired by recently-demonstrated capabilities of large (vision) language models (LLMs, VLMs) in a variety of challenging text and image tasks~\citep{openai2023gpt4v,team2023gemini}.
We build on a large body of work that leverages foundation models for decision-making in robotics~\citep{saycan,progprompt,code_as_policy,huang2022inner,curtis2024trustproc3ssolvinglonghorizon,huang2023voxposer,huang2024rekep,hu2023lookleapunveilingpower,duan2024manipulateanythingautomatingrealworldrobots,yang2024guidinglonghorizontaskmotion,keisuke} (see~\citet{hu2023robofm} for a recent survey).
However, these approaches operate under a number of different restrictive assumptions.
Some are confined to a particular task distribution (e.g., only pick-and-place) or seek to synthesize new relatively short-horizon skills (e.g., pouring liquid into a cup)~\citep{huang2023voxposer,huang2024rekep,duan2024manipulateanythingautomatingrealworldrobots}.
Others assume skills are provided, and attempt to use foundation models to compose them to solve longer-horizon tasks~\citep{saycan,progprompt,code_as_policy,huang2022inner,curtis2024trustproc3ssolvinglonghorizon,hu2023lookleapunveilingpower,quartey2024verifiablyfollowingcomplexrobot,kumar2024openworld}.
We similarly assume skills are provided, but leverage foundation models to construct abstractions that are used by a search-based planning system at inference time.
We treat the work of Hu et al.~\citep{hu2023lookleapunveilingpower} as representative of previous foundation model planning approaches and compare directly to variants in our experiments (Section~\ref{sec:experiments}).

\begin{figure*}[t]
\centering
\includegraphics[width=\textwidth]{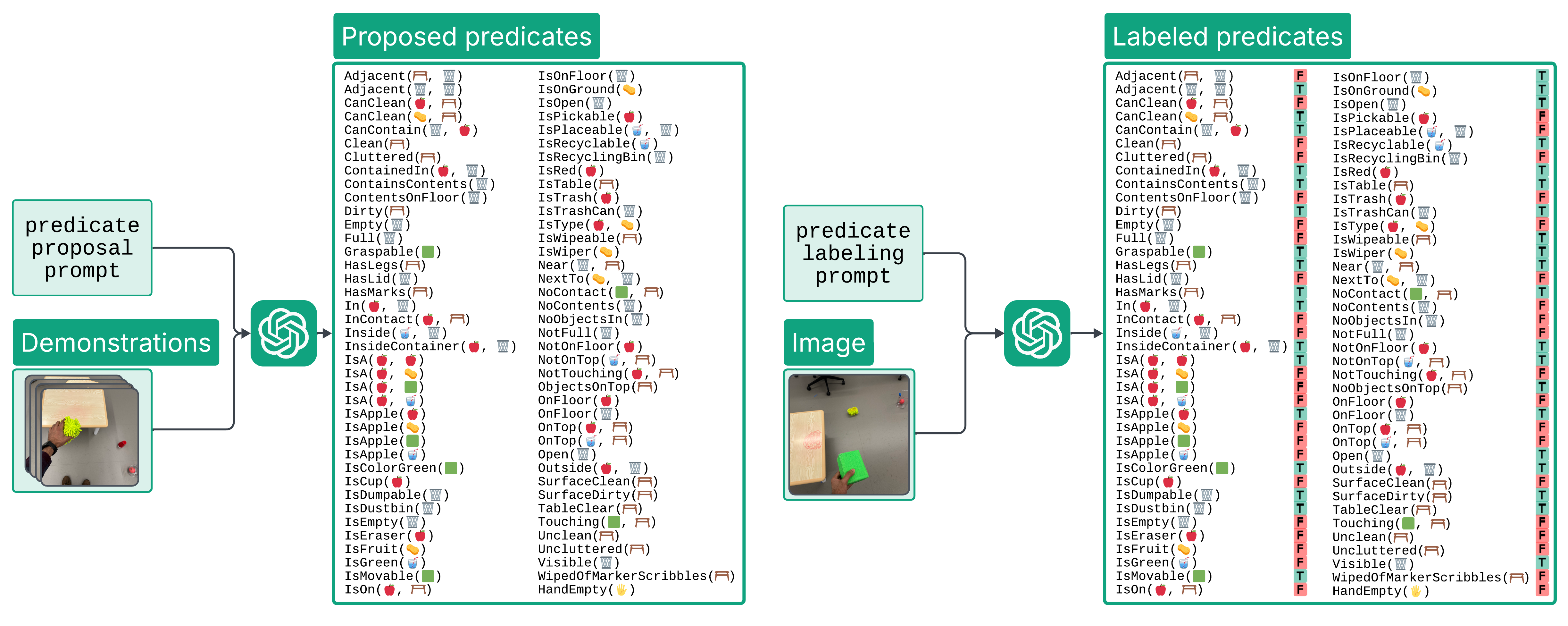}
\caption{\textbf{Labeling and proposal} (Left) A small subset of the proposed predicates for the Cleanup domain. (Right) Truth values for these predicates (green is true and red is false) in a particular state, as determined by the VLM. This data is ultimately used to subselect the correct predicates.}
\vspace{-3mm}
\label{fig:labeling-proposal}
\end{figure*}

\textbf{Learning Abstractions for Planning.} We build on a long line of work that learns abstractions for efficient planning~\citep{bertsekas1988adaptive,jong2005state,abel2016near,ugur2015bottom,asai2018classical,konidaris2018skills,silver2021learning,silver2022learning,james2022autonomous,migimatsu2022grounding,chitnis2021learning,kumar2023learning,silver2023predicate,han2024interpretinteractivepredicatelearning,liu2024BLADE,shah2024reals}.
Several of these methods assume that predicates or operators are already given~\citep{chitnis2020glib,migimatsu2022grounding,mao2022pdsketch,siegel2024language} or focus on the problem of learning predicates and other abstractions from online interaction~\citep{james2022autonomous,liang2024visualpredicatorlearningabstractworld}, or assume access to dense natural language descriptions with each demonstration~\citep{liu2024BLADE,han2024interpretinteractivepredicatelearning}.
By contrast, we aim to invent predicates and operators in an offline fashion from image-based demonstrations with no supervision on the kinds or number of predicates to learn.

Our approach takes strong inspiration from ``skills to symbols'' \citep{konidaris2018skills} and follow-up work \citep{james2022autonomous}, which also aim to learn symbolic world models from a set of skills.
James et al.~\citep{james2022autonomous} in particular learns abstractions by estimating option initiation sets and effects from interaction data, partitioning options into subgoal types, and merging objects into effect-equivalence classes to form predicates.
In contrast, our method differs in two key ways.
First, we leverage a vision--language model to propose and ground predicates directly from raw demonstrations, rather than inducing them through state partitioning.
This enables learning from just a few short, embodiment-agnostic demonstrations, while biasing the representation toward human-interpretable, ``commonsense'' predicates (e.g., \texttt{IsEraser()}, \texttt{IsColorGreen}).
Second, instead of relying on effect-equivalence criteria for learning predicates, we explicitly optimize predicate selection with a planning-efficiency objective, ensuring that the learned world models are directly tailored for effective long-horizon planning.


Our work builds directly on the bilevel planning framework of Silver et al.~\citep{silver2023predicate}, which learns predicates, operators, and samplers compatible with Task and Motion Planning (TAMP)~\citep{garrett2021integrated}.
However, unlike Silver et al., which assumes a handcrafted, noiseless object-centric state space, our approach introduces algorithmic modifications (e.g., soft precondition intersection) to handle the noise inherent in VLM-generated visual predicates.
Closely related concurrent works, \textit{VisualPredicator}~\citep{liang2024visualpredicatorlearningabstractworld} and \textit{SkillWrapper}~\citep{ramanskillwrapper}, also leverage VLMs for predicate invention \added{but are unable to learn from offline embodiment-agnostic demonstrations.
In particular, these approaches typically require the agent to execute exploratory skills and learn from contrastive signals between successful and failed skill or plan executions.
Our method targets the \emph{learning-from-demonstration} setting: it learns entirely from a small set of offline, embodiment-agnostic demonstrations (e.g., human videos) that contain only positive skill executions.
This allows us to acquire models without requiring online interaction or environment resets, enabling us to validate our approach on real-world tasks with a Boston Dynamics Spot robot (Section~\ref{sec:experiments}) trained using human video demonstrations.
Ultimately, we view our method as complementary to these online approaches~\citep{liang2024visualpredicatorlearningabstractworld, ramanskillwrapper}: \emph{pix2pred} could be used to initialize a world model from offline data, which is subsequently refined into a provably sound and complete model online by these methods.}



\section{Background and Problem Setting}
\label{sec:problem}
In this section, we describe how we model states, actions, and objects, and then provide background on predicates and symbolic world models.
We also describe the inputs and outputs of the \textit{training} (learning) and \textit{inference} (deployment) phases of our method.

\subsection{States, Actions, and Objects}
\label{subsec:object-designation}
We wish to perform decision making given an image-based state $s_t$ at timestep $t$ given a set of user-specified \textit{objects} $\mathcal{O}$.
Each object $o \in \mathcal{O}$ consists of a name, an optional type (denoted $\type$), and a descriptor string.
If a type is specified, it is associated with a continuous feature vector composed of object properties (e.g. \texttt{pose}, \texttt{size}).
Following previous work~\citep{silver2023predicate,kumar2024practice}, we assume that the agent is equipped with feature detectors (e.g. pose estimation within a pre-specified map) corresponding to each of these properties.
Descriptions are short user-provided natural-language phrases (such as ``blue mop" or ``small round coffee table") that, in the context of a particular set of images, suffice to designate a specific object.
Importantly, we assume throughout that these descriptors enable clear disambiguation and identification of the particular object within the image-based state, and the corresponding objects are present in the initial state.

The state $s_t$ itself is comprised of a set of images: $s^{\text{img}}_t = \{i^{0}_t, i^{1}_t, \ldots, i^{j}_t\}$ (potentially captured from multiple cameras) and an ``object-centric'' state $s^{\text{obj}}_t$, which includes the set of continuous feature vectors (e.g. \texttt{pose}, \texttt{size}) for every unique \textit{object} with corresponding type that is in the images.
Overall, $s_t \gets s^{\text{img}}_t \mathbin\Vert s^{\text{obj}}_t$, where $\mathbin\Vert$ denotes concatenation. Thus, the overall state space $S$ is composed of an image-based state space, and a feature-based state space.

$A$ is the agent's action space defined by a finite set of \emph{skills} $\C$ that are temporally-extended low-level robot behaviors.
Each skill $C((\type_1, \dots, \type_v), \Theta) \in \C$ has a semantically-meaningful name, a policy function, optional discrete typed object parameters $(\type_1, \ldots, \type_v)$, and optional continuous parameters $\Theta$.
For instance, a skill \texttt{Grasp} for picking up an object might have one discrete parameter of type \texttt{object} and a 6-dimensional $\Theta$ that is a placeholder for a specific grasp.
An \emph{action} $a \in \A$ is a skill $C \in \C$ with discrete and continuous arguments fully specified: $a = C((o_1, \ldots o_v), \theta)$.
An action executes an underlying policy from the policy function (e.g. a learned grasp policy) to modify the state.

\begin{figure*}[t]
\centering
\includegraphics[width=\textwidth]{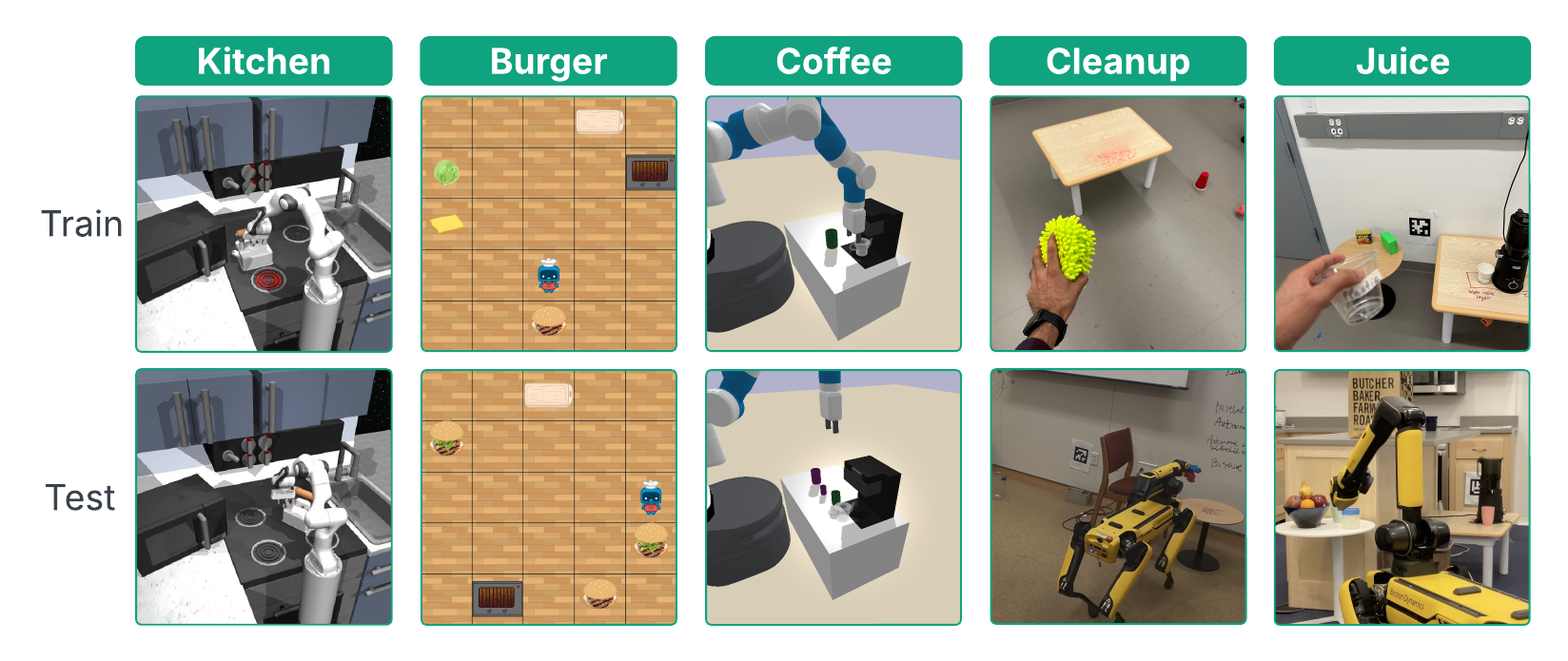}
\caption{\textbf{Domains.}
Top row: train task example illustrations. Bottom row: test task example illustrations.
}
\vspace{-1em}
\label{fig:environments}
\end{figure*}

\subsection{Predicates, Symbolic World Models, and Planning}
\label{subsec:predicates-background}
A predicate is characterized by a name, an ordered list of $m$ arguments $(\type_1, \dots, \type_m)$ (note that the types of the arguments are not necessarily unique). and a classifier function $c_\psi : S \times \Lambda^m \to \{\text{true}, \text{false}\}$ parameterized by the arguments.
A \emph{ground atom} $\ground{\psi}$ consists of a predicate $\psi$ and objects $(o_1, \dots, o_m)$.
Each ground atom induces a particular binary state classifier $c_{\ground{\psi}} : S \to \{\text{true}, \text{false}\}$, where $c_{\ground{\psi}}(x) \triangleq c_{\psi}(x, (o_1, \dots, o_m))$.
For example, the ground atom \texttt{Holding(spot, apple)} would have the value ``true'' if the \texttt{apple} is being held by the \texttt{spot}.
In this work, we distinguish between \textit{feature-based} predicates that operate exclusively over the object-centric state, and \textit{visual} predicates that operate on the image-based state.

A set of predicates $\Psi$ induce an abstract state space $S^{\Psi}$ of the underlying continuous state space $S$.
In order to enable planning, we require an action model $\Omega$ that specifies an abstract transition model over $S^{\Psi}$.
We learn such an action model in the form of symbolic PDDL-style~\citep{McDermott1998PDDLthePD,fox2003pddl2} operators that are each linked to a particular skill from the domain's action space.
In domains where skills have continuous parameters, each operator is associated with a generative \textit{sampler} that proposes continuous parameters for the associated skill~\citep{chitnis2021learning}.

\textbf{Planning.} Together, the predicates and operators specify a symbolic abstract world model that can be used in conjunction with samplers for efficient hierarchical planning such as Task and Motion Planning (TAMP)~\citep{srivastava2014combined,garrett2021integrated}.
In this work, we choose to use a simpler planning procedure that does not require a precise simulator or transition model~\citep{kumar2024practice}.
At a high level, our planning procedure takes in the initial state $s_0$ and first evaluates all the provided predicates $\Psi$ to yield an \textit{abstract state}.
Then, it calls a PDDL planner to achieve the goal from this abstract state.
We can simply execute each action of this plan directly and check whether the goal is achieved at the end (which we do in all our simulation domains), or execute actions one-by-one and replan from resulting states as necessary (which we demonstrate in some of our real-world deployment).
The precise details of our planning procedure are not necessary to understand what follows; we defer them to Appendix~\ref{appendix:planner}.

\subsection{Learning and Deployment}
\label{subsec:training-time}
\textbf{Learning Phase.}
We learn from a set of $n$ \textit{demonstrations} $\mathcal{D}$, starting with a set of initial predicates $\Psi^{\text{init}}$.
Each demonstration $d$ consists of (1) a set of objects $\mathcal{O}^d$, (2) a k-step trajectory $\tau^d = [s_0, a_0, s_1, a_1, \ldots, s_k]$ of states and actions, and (3) a conjunctive \textit{goal expression} $g^d$ that holds true only in the final state $s_k$.
Here, $g^d$ is a conjunction of ground atoms with predicates from $\Psi^{\text{init}}$, and objects from the object set $\mathcal{O}^d$.
The initial predicate set $\Psi^{\text{init}}$ consists of predicates that are necessary to specify the demonstration goals (we denote this subset $\Psi^{G}$), and optionally some additional predicates that might be known ahead of time.
As a concrete example, for the Cleanup domain in Figure~\ref{fig:cleanup-domain-overview}, the initial predicate set consists of \{\texttt{WipedOffTable(?obj)}, \texttt{InsideContainer(?obj1, ?obj2)}, \texttt{Holding(?robot, ?obj)}, \texttt{HandEmpty(?robot)}\}, the first demonstration has the goal \texttt{WipedOffTable(play\_table)}, and the second has the goal \texttt{InsideContainer(green\_block, cardboard\_box)}.

\textbf{Deployment Phase.}
Given (1) a new set of objects $\mathcal{O}^{\text{test}}$ with the same types as seen in the demonstrations, (2) a novel initial state $s_0$, and (3) a novel goal $g^{\text{test}}$ expressed using $\Psi^{\text{init}}$, we must output a sequence of $m$ actions (which we call a \textit{plan}) that achieves some new state $s_m$ where $g^{\text{test}}$ holds.
We refer to the tuple of objects, initial state, and goal as a \textit{task}.


\section{Learning Symbolic World Models}
\label{sec:method}
Given a demonstration set $\mathcal{D}$ (Section~\ref{subsec:training-time}), we learn a PDDL-style~\citep{McDermott1998PDDLthePD,fox2003pddl2} symbolic world model $\mathcal{W}^{\mathcal{D}}$ by inventing additional predicates $\Psi^{\mathcal{D}}$, learning operators defined by both the newly-invented predicates and the initial predicates $\Psi^{\mathcal{D}} \cup \Psi^{\text{init}}$, and, where applicable, training generative samplers for continuous skill parameters~\citep{chitnis2021learning}. We adapt prior work for sampler learning (Appendix~\ref{appendix:sampler-learning}) and focus on inventing visual predicates and operators from demonstrations.

\subsection{Proposing an Initial Pool of Visual Predicates}
\label{subsec:vlm-proposal}
The first step of our approach is to generate a pool of candidate visual predicate names by prompting a VLM on each demonstration $d \in \D$.
For a demonstration $d \in \D$ of length $k = len(d)$, we extract the image-based state at each timestep $s^{\text{img}}_{t}, t \in [0, k-1]$ (i.e., the states before and after each skill execution).
We add a text heading to each image corresponding to which timestep $t$ in the demonstration it belongs to.
We then pass all these images in sequence, along with description of the actions (name and specific arguments) executed in between them ($a_{0}, a_1, \ldots, a_{k-1}$), and the descriptors of the objects in each demonstration, directly to a VLM.
We prompt the VLM to output a set of proposals (in the form of natural language strings) for ground atoms based on these inputs.

Given a set of ground atoms proposed for each demonstration, we then automatically parse this set to discover atoms that are syntactically incorrect.
In particular, we remove atoms that include object names that are not part of the demonstration $d$ (for instance, the VLM might propose an atom \texttt{Inside(apple, room1)}, though there is no object named \texttt{room1}).
\added{We then deterministically \textit{lift} each of these atoms into a visual predicate by creating typed variables for each object (for instance, an atom \texttt{OnTop(apple, short\_round\_coffee\_table)} would be lifted to \texttt{OnTop(?m: movable, ?t: table)}, where \texttt{movable} and \texttt{table} are input types specified as part of specifying the objects as defining the problem and environment)}.
Finally, we remove any duplicate lifted atoms and add only unique predicates to our pool.
We obtain a final pool of predicates $\Psi_{\text{pool}}^{\text{vis}}$ by running this procedure on each demonstration.

\subsection{Implementing Visual Predicates with a VLM}
\label{subsec:vlm-predicates}
We implement the classifier function $c_\psi$ for all visual predicates $\psi^{\text{vis}}$ as a query to a VLM.
Specifically, we create a prompt string $\text{txt}_{\psi}$ with each predicate's name and arguments (e.g. \texttt{NoObjectsOnTopTable(?surface)}).
When the predicate is ground, we substitute the descriptors of the specific objects used to ground it into the prompt (e.g. \texttt{NoObjectsOnTopTable (short\_round\_coffee\_table)}).
To evaluate the predicates, we pass the prompt with these ground atom strings into a VLM with additional instructions, asking it to output either ``true", ``false", or ``unknown" (which we take to be equivalent to ``false") for each predicate.
If the current state is not the initial state (i.e., for $s_t: t > 0$) of a task, we also provide the previous state's image-based state, ground atom labeling, and executed action.
We discuss our approach to prompting for atom labeling and proposal in detail and provide examples of our prompts in Appendix~\ref{appendix:prompting}.

\subsection{Learning Symbolic World Models via Optimization}
\label{subsec:pred-selection}

Algorithm~\ref{alg:hill-climbing} shows the pseudocode of our hill-climbing procedure for learning predicates and operators.
Here, $\Psi^{\text{vis}}_{\text{pool}}$ is generated by the procedure described in Section~\ref{subsec:vlm-proposal}.
Where possible (such as in simulation domains where object features are available), we combine these with additional ``feature-based'' predicates that are generated from a programmatic grammar over the object-centric state space, as described in~\citet{silver2023predicate}), to create an overall pool $\Psi_{\text{pool}} \gets \Psi_{\text{pool}}^{\text{vis}} \cup \Psi_{\text{pool}}^{\text{feat}}$.
\added{The feature-based predicates are intended to capture relations that are challenging to capture visually (e.g. relations involving robot joint angles).}
$\Psi_{\text{pool}}$ is passed into the hill-climbing procedure from~\citep{silver2023predicate} that optimizes a planning-based objective ($J(\ldots)$). Internally, it performs operator learning (OperatorLearn) and outputs operators in terms of a small subset of predicates from $\Psi_{\text{pool}}$.

\begin{algorithm}[h]
\caption{Model Learning}
\label{alg:hill-climbing}
\small
\KwIn{Demos $\mathcal{D}$, initial predicates $\Psi_{\text{init}}$}
$\Psi_{\text{pool}} \gets \Psi_{\text{init}} \cup \Psi_{\text{pool}}^{\text{vis}} \cup \Psi_{\text{pool}}^{\text{feat}}$\;
$\Psi_{\text{selected}} \gets \emptyset$, $J_{\text{prev}},J_{\text{curr}} \gets \infty$\;
\While{$J_{\text{prev}} - J_{\text{curr}} > J_{\text{thresh}}$}{
    $J_{\text{prev}} \gets J_{\text{curr}}$\;
    $\psi^* \gets \arg\min_{\psi \in \Psi_{\text{pool}}} J(\mathcal{D}, \text{OperatorLearn}(\mathcal{D}, \Psi_{\text{selected}} \cup \{\psi\}), \Psi_{\text{selected}} \cup \{\psi\})$\;
    $J_{\text{curr}} \gets J(\mathcal{D}, \text{OperatorLearn}(\mathcal{D}, \Psi_{\text{selected}} \cup \{\psi^*\}), \Psi_{\text{selected}} \cup \{\psi^*\})$\;
    $\Psi_{\text{selected}} \gets \Psi_{\text{selected}} \cup \{\psi^*\}$, $\Psi_{\text{pool}} \gets \Psi_{\text{pool}} \setminus \{\psi^*\}$\;
}
\Return predicates $\Psi_{\text{selected}} \cup \Psi^{G}$, $\text{OperatorLearn}(\mathcal{D}, \Psi_{\text{selected}})$
\end{algorithm}

We make \added{a few core modifications to the version of Algorithm~\ref{alg:hill-climbing} from~\citep{silver2023predicate}. We regularize the optimization by early-stopping the hill-climbing optimization via a hyperparameter ($J_{\text{thresh}}$) that we tune on a small validation set.
Noise also tends to adversely impact operator learning: small differences in atom values before and after a particular skill is executed lead to learning many operators that are each overfit to specific cases, instead of one general operator that handles many transitions.
We combat this by modifying the precondition induction procedure of operator learning to be more noise aware, as well as simply deleting operators that model less than a certain fraction of the data (controlled by a hyperparameter).}
See Appendix~\ref{appendix:op-learning} for more details.

A key advantage of our approach is that the optimization will automatically select among several closely-related and even synonymous predicates (such as \texttt{NextTo} vs. \texttt{Touching} in Figure~\ref{fig:labeling-proposal}) to find the predicates that are most accurately labeled and also important for planning across the training demonstrations.

\section{Experiments}
\label{sec:experiments}
We design experiments to answer the following questions.

\begin{tightlist}
\item[\hspace{2em}\textbf{Q1}.] How well does \ours{} generalize to novel,  more complex goals, especially when compared to an imitation approach that doesn't use a planner?
\item[\hspace{2em}\textbf{Q2}.] How critical is it to perform explicit score-based optimization for subselection of predicates?
\end{tightlist}

\textbf{Domains.} We now describe our experimental domains and tasks \added{(illustrated in Figure~\ref{fig:environments})}, with details in Appendix~\ref{appendix:additional-experiments}. We implement 2 real-world tasks on a Boston Dynamics Spot robot in different real-world buildings.
For the purposes of extensive comparison against baselines, and to demonstrate the broad applicability of our method across sequential decision-making problems, we use 3 simulated domains that were introduced by previous work and lightly adapted for our setting.
\begin{tightlist}
    \item \textit{Cleanup:} A real-world Spot clears objects into a bin (sometimes dumping the bin) and wipes a tabletop. Testing tasks vary objects, backgrounds, and require more complex compositions of skills.
    \item \textit{Juice:} A real-world Spot operates a cold-press juicer by setting up cups, loading fruit, and running the machine. Testing tasks introduce new objects, backgrounds, and goals requiring demonstration composition.
    \item \textit{Kitchen:} Originally introduced by~\citet{gupta2019relaypolicylearningsolving}: a robot must boil water by placing a kettle on a stove burner and turning it on. Testing tasks involve using a different burner.
    \item \textit{Burger:} A simulated burger-making grid world~\citep{wang2023demo2code}. We test three scenarios: \textit{Bigger Burger} tasks require stacking extra patties, \textit{More Burger Stacks} tasks involve assembling multiple burgers instead of one, and \textit{Combo Burger} tasks demand combining multiple different ingredients at test time without seeing combining at training time. Test tasks involve sequences of up to 30 skill executions
    \item \textit{Coffee:} A simulated task~\citep{silver2022learning} where a robot dispenses coffee into a jug from a coffee machine and pours coffee into cups of varying sizes. Test tasks add more cups and novel object poses.
\end{tightlist}

\begin{figure*}[t]
\includegraphics[width=\textwidth]{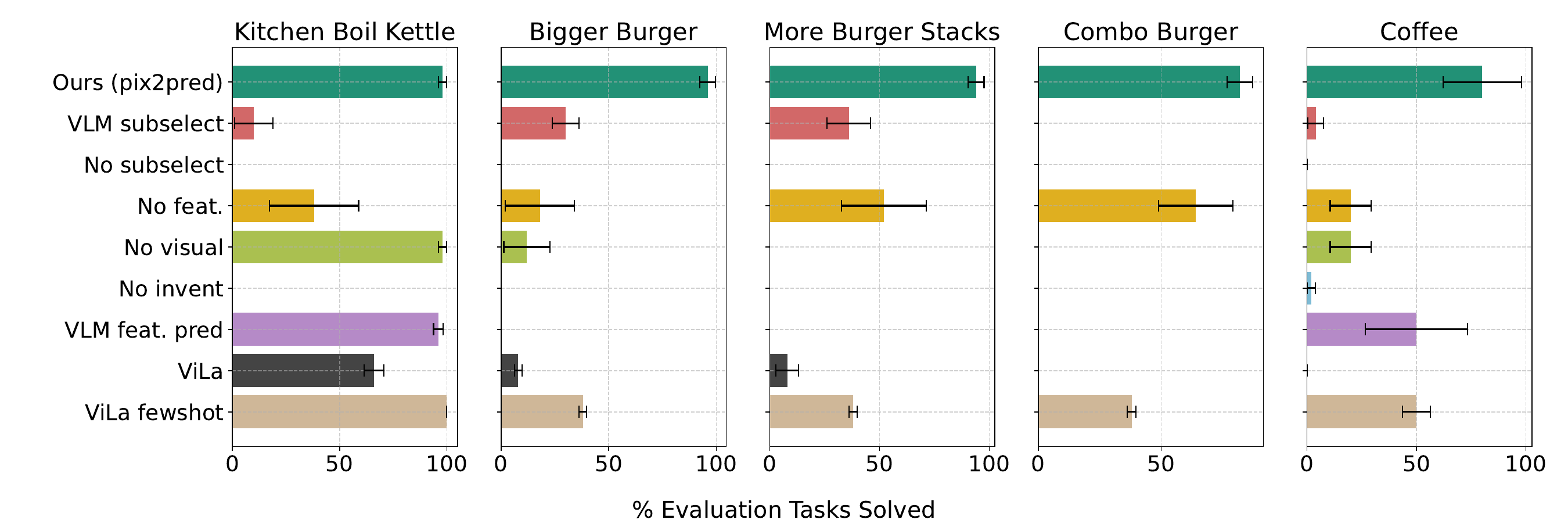}
\centering
\caption{\textbf{\ours{} versus baselines in simulation}. Percent of evaluation tasks solved across all our simulated domains. All results are averaged over 5 seeds. Black bars denote standard deviations.
}
\label{fig:burgerbarplot}
\vspace{-1em}
\end{figure*}

\textbf{Approaches.} We evaluate the following ablations and baselines in all simulated domains.
\begin{tightlist} \item \textit{Ours}: Our full main method: \ours{}.
\item \textit{VLM subselect}: Ablation where the VLM directly selects a compact set of visual predicates ($\Psi^{\text{vis}}$); feature-based predicates ($\Psi^{\text{feat}}$) are still selected via hill-climbing.
\item \textit{No subselect}: Ablation with no hill-climbing; selects all predicates in $\Psi_{\text{pool}}$. \item \textit{No invent}: Ablation with no new predicates beyond $\Psi_{\text{init}}$.
\item \textit{No visual}~\citep{silver2023predicate}: Ablation inventing only feature-based predicates ($\Psi^{\text{feat}}$), equivalent to.
\item \textit{No feature}: Ablation inventing only visual predicates.
\item \textit{VLM feat. pred}~\citep{hu2023lookleapunveilingpower}: Baseline where a VLM is used to directly output predicates expressed as arbitrary code functions over $S^{\text{obj}}$~\citep{han2024interpretinteractivepredicatelearning}.
\item \textit{ViLa}: Baseline where a VLM plans directly without abstractions; we test \textit{ViLa-pure} (original version) and \textit{ViLa-fewshot} (novel variant that adds demonstrations to the few-shot prompt).
\end{tightlist}

\textbf{Experimental Setup.}
In simulation, we vary object counts and locations across $5$ seeds, sample $10$ test tasks per seed, and allow one plan execution per task. In real-world domains, all demonstrations are collected from the point-of-view of a human, no object-centric state is available in the training data, and learned world models are evaluated on a robot. We compare our method to ViLA-fewshot using a fixed training set ($6$ demos for Cleanup, $10$ for Juice) and measure success as whether a correct plan is generated from a given initial state across $5$ generalization-focused test tasks. GPT-4o~\citep{openai2023gpt4v} is used as the VLM in simulation, and Gemini-2.0-flash~\citep{team2023gemini} in the real world. All approaches use the same set of underlying skills in each domain.
Specific details, including the precise prompts, skill implementations, and hyperparameters used can be found in Appendix~\ref{appendix:additional-experiments}.

\renewcommand{\arraystretch}{1.4}
\begin{table}[t]
\vspace{1em}
\centering
\setlength{\tabcolsep}{3pt}
\begin{tabular}{llccccc}
\toprule
\textbf{Task} & \textbf{Approach} & \textbf{New} & \textbf{New} & \textbf{More} & \textbf{Novel} & \textbf{Novel} \\
             &                    & \textbf{obj.} & \textbf{vis.} & \textbf{obj.} & \textbf{goal 1} & \textbf{goal 2} \\
\midrule
\multirow{2}{*}{Cleanup}
  & \textsc{Ours}         & 3/3 & 3/3 & 3/3 & 3/3 & 3/3 \\
  & ViLA-fewshot          & 3/3 & 3/3 & 2/3  & 0/3   & 3/3  \\
\midrule
\multirow{2}{*}{Juice}
  & \textsc{Ours}         & 2/3  & 2/3  & 3/3 & 2/3  & 1/3  \\
  & ViLA-fewshot          & 0/3   & 2/3  & 3/3 & 0/3   & 0/3   \\
\bottomrule
\end{tabular}
\vspace{1em}
\caption{\textbf{Real-world generalization results.} We evaluate on 5 different tasks testing different aspects of generalization in our 2 real world domains after training on the exact same dataset. Success out of 3 random initial states is reported. Additional details on task construction and evaluation are in Appendix~\ref{appendix:additional-experiments}.}
\label{table:real-world-results}
\end{table}

\textbf{Results and Analysis.}
Figure~\ref{fig:burgerbarplot} shows success rates in simulated domains.
\ours{} outperforms prior approaches on 4 of 5 domains tested, often by a wide margin, and generalizes far better to complex tasks than ViLa few-shot (\textbf{Q1}).
On the short-horizon kitchen task (2 actions), ViLa with demonstrations performs comparably to \ours{}, but its performance drops sharply on longer-horizon burger tasks (15+ actions).
We find that ViLa tends to pattern-match training demonstrations and struggles with test tasks requiring true generalization—for example, handling initial states where the agent starts out holding an object (seen in $\sim$50\% of Burger test tasks but never in training).
ViLa fails to put down objects before picking up new ones, while \ours{} invents a \texttt{HandEmpty} predicate that enables the planner to handle such cases.

Our method also outperforms ViLA in real-world domains (Table~\ref{table:real-world-results}).
ViLA handles new backgrounds and object instances and quantities to some extent but struggles with novel goals requiring composition across demonstrations, especially in the complex Juice domain.
In contrast, \ours{} generalizes across object instances, visual backgrounds, object counts, and novel goal compositions with just $6$ and $10$ human-collected demonstrations.
Qualitatively, \ours{} invents few but meaningful predicates (e.g., \texttt{OnFloor} and \texttt{IsEraser} in Cleanup; \texttt{JuiceMachineOpen}, \texttt{InsideWasteValveRegion}, and \texttt{NearJuiceValve} in Juice).
See Appendix~\ref{appendix:examples-from-learning} for additional details.

We also find that explicit predicate subselection significantly improves test performance over direct VLM predicate selection (\textbf{Q2}).
The VLM subselect baseline consistently underperforms \ours{} across all four task distributions.
Qualitatively, VLM subselect picks many unnecessary predicates. In the Burger domain, for instance, it selects useful concepts like \texttt{Cooked(?p1:patty)} and \texttt{Chopped(?l1:lettuce)}, but also inconsistent and redundant ones like \texttt{Raw(?p1:patty)}, \texttt{Available(?p1:patty)}, and \texttt{Whole(?l1:lettuce)}.
This bloats the operator set, causing overfitting and poor generalization.
Finally, the ``no subselect'' baseline fails entirely, showing that subselection is critical.
In most of our domains, the VLM proposes over $100$ distinct predicates; trying to learn operators over these directly leads to severe overfitting.

There are two possible causes of failure for our approach: (i) \ours{} learns a correct model, but fails to make a correct plan in test tasks because the VLM incorrectly classifies one or more predicates at test-time, and (ii) \ours{} fails to learn a model that generalizes to planning on test tasks (despite predicate values being ``correct'').
Upon manual inspection we found that all failures in our real-world environments, and the majority of failures in simulation environments, are due to (i).


\section{Conclusion}
\label{sec:conclusion}
We proposed \ours{}, a method for inventing symbolic world models capable of enabling planning directly from input pixels.
We found that our approach, from just a handful of demonstrations (3-12), is able to invent semantically meaningful predicates that afford efficient planning and aggressive generalization to novel object instances, numbers, task horizons, and goals across a range of problems.
\section{Limitations}
\label{sec:limitations}

There are several important limitations of \ours{} in its current form.
First, a key assumption of our approach is that, for each new task, all relevant objects in the scene are specified with unambiguous descriptors. This is currently done manually for the real-world environments and can be time-consuming.
A second assumption is the full observability of objects in all states. These assumptions do not hold in all tasks of interest, especially when deploying in novel real-world environments. Future work could attempt to extract descriptors automatically from foundation models via a form of automatic prompt optimization~\citep{khattab2024dspy}, and extend the core ideas of \ours{} to partially observable settings.

A further limitation lies in the hill-climbing optimization procedure (Section~\ref{subsec:pred-selection}) at the core of our model learning approach: it can be extremely slow --- taking many seconds to evaluate the objective value of a single candidate predicate set --- especially as the number of demonstrations and size of the initial predicate pool grow. We also found it to be quite sensitive to the choice of hyperparameters. Future work could develop more efficient and noise-tolerant optimization algorithms for predicate selection, as well as improved objectives for them. Relatedly, operator learning algorithms could be improved to better handle large input predicate sets and noisy predicates.

Finally, our approach assumes input demonstrations are segmented in terms of a provided set of parameterized skills with names that correspond to their function. These skills, as well as their underlying policies, are assumed to be provided ahead of time. Future work could integrate \ours{} with low-level skill-learning approaches, enabling learning from very low-level robot demonstrations or potentially even from large-scale internet video corpora.

Overall, we hope our work inspires future progress toward learning structured and interpretable world models that enable efficient inference in and strong generalization across complex robotic manipulation problems.
\newpage
\bibliography{aaai25}
\clearpage
\appendix
\section{Appendix}

\subsection{Symbolic Operators}
\label{appendix:ops}
Given a particular state $s_{t}$, as well as predicates $\Psi$, we can compute the corresponding \textit{abstract state} (denoted by $s^{\Psi}_{t}$) by selecting all the ground atoms that are ``true" in $s_{t}$. 
Formally, $s^{\Psi}_t = \textsc{Abstract}(s_t, \Psi) \triangleq \{ \ground{\psi} : c_{\ground{\psi}}(s_t) = \text{true}, \forall \psi \in \Psi \}.$
Then, each operator $\operator$ is a tuple $\operator = \langle \oparguments, \preconditions, \addeffects, \deleteeffects, \C \rangle$. 
Here, $\oparguments$ are typed variables representing the operator's arguments, $\preconditions$, $\addeffects$ and $\deleteeffects$ are sets of predicates representing operator preconditions, add effects and delete effects respectively, and $\C$ is a controller associated with the operator. 
Note importantly that the discrete parameters of $\C$ must be a subset of $\oparguments$. 
Specifying objects of the correct type as arguments induces a \emph{ground operator} $\ground{\operator} = \langle \ground{\preconditions}, \ground{\addeffects}, \ground{\deleteeffects}, \ground{\C} \rangle$.
Ground operators define a (partial) abstract transition function: given $\ground{\operator} = \langle \ground{\preconditions}, \ground{\addeffects}, \ground{\deleteeffects}, \ground{\C} \rangle$, if $\ground{\preconditions} \subseteq s$, then the successor abstract state $s'$ is
$(s \setminus \ground{\deleteeffects}) \cup \ground{\addeffects}$ where $\setminus$ is set difference.

A ground operator specifies a corresponding ground controller $\ground{\C}$.
To extract an action $a$ from this ground controller, we must specify values for the parameters $\Theta$ associated with controller $\C$.
Following previous work~\citep{silver2023predicate,kumar2023learning,chitnis2021learning}, we leverage continuous parameter \textit{samplers} to do this.

\subsection{Planner Implementation Details}
\label{appendix:planner}
\begin{algorithm}[h]
    \DontPrintSemicolon
    \nl \textbf{Input:} Task $\langle \O, s_0, g \rangle$, predicates $\Psi$, operators $\Omega$, samplers $\Sigma$.\;
    \nl $s^{\Psi}_0 \gets \textsc{Abstract}(s_0, \Psi)$\;
    \nl Call classical planner to generate an abstract plan:
    $\text{plan}^{\text{abs}} = \ground{\operator_0}, \ground{\operator_1}, ..., \ground{\operator_{m}} \gets \text{Planner}(s^{\Psi}_0, \O, \Psi, \Omega, g)$\;
    \nl Extract controller sequence with discrete args filled in: 
    $\text{skeleton} = \ground{\C_0}, \ground{\C_1}, ..., \ground{\C_{m}}$\;
    \nl For $i=0, \dots, m$:\;
    \nl \quad Sample $\theta \sim \sigma_{\operator_i}$, where $\theta$ represents the continuous parameters of $\C_i$ and $\sigma_{\operator_i}$ is the sampler associated with $\operator_i$\;
    \nl \quad Use $\theta$ to fully ground controller $\ground{\C_{i}}$ into an action $a_i$.\;
    \nl \textbf{execute} action $a_i$ and obtain the following state $s_{i + 1} \gets f(s_i, a_i)$\;
\caption{Planning and Execution pseudocode}
\label{alg:planning-to-solve-tasks}
\end{algorithm}

Algorithm~\ref{alg:planning-to-solve-tasks} shows the pseudocode for the planning and execution strategy we implement following recent work~\citep{kumar2024practice}. Given an initial state $x_0$ for a task, we simply evaluate the classifiers of the Initial predicates $\Psi$ to convert the state into an abstract state $s_0$. We then call a classical planner with the task object set $\O$ and operators $\Omega$ to compute an abstract plan that achieves the goal (if one exists). We extract the ground controller sequence for this abstract plan, and then simply greedily execute each controller sequentially by calling the sampler associated with each operator.

We note that more sophisticated planning strategies are possible. In particular, provided a photorealistic simulation environment for the transition function $f$, we could leverage many task and motion planning~\citep{garrett2021integrated}, such as bilevel planning~\citep{silver2023predicate,kumar2023learning,chitnis2021learning}.

\subsection{Operator Learning}
\label{appendix:op-learning}
We adapt the ``cluster and intersect'' operator learning strategy from previous work~\citep{chitnis2021learning} to handle noise in predicate values inherent to our setting.
Specifically, we learn a set of operators $\Omega$ from our demonstrations $\D$ and predicates $\Psi$ in four steps. Of these, the first three steps are largely taken directly from previous work: we introduce a modification to the third step, as well as the final step to handle noise. 
\begin{enumerate}
    \item \textit{Partitioning}: Each demonstration can be expressed as a sequence of transitions $\{(s, a, s')\}$, with $s, s' \in \S$ and $a \in \A$.
    Recall that each action $a$ is a controller with particular discrete and continuous arguments specified; let $\ground{\C}$ denote the corresponding controller with the same discrete object argument values, but continuous parameter values left unspecified.
    First, we use $\Psi$ to \textsc{Abstract} all states $s, s'$ in the demonstrations $\D$, creating a dataset of transitions $\{(s^{\Psi}, a, s'^{\Psi})\}$ with $s^{\Psi}, s'^{\Psi} \in \S^\Psi$.
    Next, we partition these transitions via the following equivalence relation: $(s_1^{\Psi}, a_1, s_1'^{\Psi}) \equiv (s_2^{\Psi}, a_2, s_2'^{\Psi})$ if the effects and controllers \emph{unify}, that is, if there exists a mapping between the objects such that $\ground{\C_1}$, $(s_1^{\Psi} - s_1'^{\Psi})$, and $(s_1'^{\Psi} - s_1^{\Psi})$ are equivalent to $\ground{\C_2}$, $(s_2^{\Psi} - s_2'^{\Psi})$, and $(s_2'^{\Psi} - s_2^{\Psi})$ respectively.
    After this step, we have effectively `clustered' all transitions in $\D$ together: we can associate each transition $\{(s^{\Psi}, a, s'^{\Psi})\}$ with a particular equivalence class.
    \item \textit{Arguments and Effects induction:} For each equivalence class created in the previous step, we create $\oparguments$ by selecting an arbitrary transition $(s^{\Psi}, a, s'^{\Psi})$ and replacing each object that appears in the controller $\ground{\C}$ or effects with a variable of the same type. 
    This further induces a substitution $\delta: \oparguments \to \O$ for the objects $\O$ in this transition. 
    Given this, the $\addeffects$, and $\deleteeffects$ can then be created by applying $\delta$ to $(s'^{\Psi} - s^{\Psi})$, and $(s^{\Psi} - s'^{\Psi})$ respectively.
    By construction, for all other transitions $\tau$ in the same equivalence class, there exists an injective substitution $\delta_\tau$ under which the controller arguments and effects are equivalent to the newly created $\addeffects$, and $\deleteeffects$.
    \item \textit{Precondition learning:} The only remaining component required to turn each equivalence class into an operator is the operator preconditions.
    For this, we perform an intersection over all abstract states in each equivalence class~\cite{bonet2019learning,curtis2021discovering}. Recall that an abstract state is simply the collection of ground atoms that are `true', thus taking an intersection amounts to finding the set of atoms that are \textit{always} true across every initial state of every transition $(s^{\Psi}, a, s'^{\Psi})$ in the equivalence class.
    However, since some of our predicate classifiers might be noisy they might not \textit{always} hold.
    Thus, we take a `soft' intersection: we take any atom as a precondition that is true across more than a specific percentage (set by a hyperparameter $h_{\text{pre\_frac}}$) of transitions in the equivalence class.
    More specifically, $\preconditions \gets \bigcap_{\tau=(s^{\Psi}, \cdot, \cdot)} \delta^{-1}_\tau(s^{\Psi})$ if $\frac{|\delta^{-1}_\tau(s^{\Psi})|}{|\D|} \geq h_{\text{pre\_frac}}$,
    where $\delta^{-1}_\tau(s^{\Psi})$ substitutes all occurrences of the objects in $s$ with the parameters in $\oparguments$ following an inversion of $\delta_\tau$, and discards any atoms involving objects that are not in the image of $\delta_\tau$.
    $|\delta^{-1}_\tau(s^{\Psi})|$ denotes the number of transitions in $\D$ in which the lifted atom $|\delta^{-1}_\tau(s^{\Psi})|$ holds, and $|\D|$ denotes the total number of transitions in $\D$. In our experiments, we set $h_{\text{pre\_frac}}$ to 0.8.
    \item \textit{Pruning low-data operators:} Noise in the atom values often causes there to be equivalence classes with very few data points, since transitions that have been affected by noise do not unify with other transitions in our dataset.
    These lead to operators that are overfit to those particular noisy transitions, which are undesirable.
    We combat this via simply discarding learned operators that have data below a certain fraction (denoted by hyperparameter $h_{\text{data\_frac}}$) of the total transitions associated with a particular controller $\C$. In particular, let $|\tau_{\C}|$ denote the number of transitions in $\D$ where the action $a$ involves a particular controller $\C$. For any operator $\omega$, let $|\tau_{\omega}|$ denote the number of datapoints associated with the equivalence class used to construct that operator. We only keep an operator $\omega$ if $\frac{|\tau_{\omega}|}{|\tau_{\C}|} \geq h_{\text{data\_frac}}$. In our experiments, we set $h_{\text{data\_frac}} = 0.05$.
\end{enumerate}

\subsection{Sampler Learning}
\label{appendix:sampler-learning}
To enable execution, we must also learn samplers for proposing continuous controller parameters (Algorithm~\ref{alg:planning-to-solve-tasks}). Adapting prior work \citep{silver2023predicate, kumar2023learning, chitnis2021learning}, we train one sampler per operator, defined as:
\[
\sampler(s, o_1, \dots, o_k) = u_\sampler(s[o_1] \oplus \cdots \oplus s[o_k]),
\]
where \(x[o]\) is the feature vector for \(o\), \(\oplus\) denotes concatenation, and \(s_\sampler\) is a learned model. Treating this as supervised learning, we use operator-specific datasets \(\tau_{\omega}\), where each of these datasets is composed of all transitions from the equivalence class used to induce this operator (Appendix~\ref{appendix:op-learning}). Each transition \((s_i^{\Psi}, a_{i+1}, s_{i+1}^{\Psi}) \in \tau_{\omega}\) maps operator arguments \(\oparguments\) to objects via \(\delta: \oparguments \to \O_{\tau}\). Using this mapping, the input for training is \(s[\delta(v_1)] \oplus \cdots \oplus s[\delta(v_k)]\), where \((v_1, \dots, v_k) = \oparguments\), and the output is the continuous parameter vector \(\theta\), which are the continuous parameters used for \(a_{i+1}\).

Each sampler is parameterized by two neural networks. The first predicts a Gaussian distribution over \(\theta\), regressing its mean and covariance. The second network classifies \((s[o_1] \oplus \cdots \oplus s[o_k], \theta)\) as valid or invalid, enabling rejection sampling from the Gaussian. Negative examples are transitions outside \(\tau_{\omega}\) but using the same controller.

\subsection{Additional Prompting Details}
\label{appendix:prompting}
Here, we provide additional details about how we prompt a VLM for both labeling and proposal of atoms.

\subsubsection{Atom Labeling.}  
Recall from Section~\ref{subsec:vlm-predicates} that the goal of atom labeling is to obtain the truth value of a particular ground atom $\ground{\psi}(o_0, \ldots, o_l)$ given an image-based state at timestep $t$ in a trajectory $s_t^{\text{img}}$.
We do this by prompting a VLM with a text prompt that includes a string representation of the atom (e.g. \texttt{Cooked(patty1)}), along with the relevant images in $s_t^{\text{img}}$, and asking it to output the truth value of the atom.
Since we usually want to query for the values of many atoms at once, we provide all atoms in one single query and ask a VLM to label their truth values in batch. Having separate queries to label each ground atom would probably lead to more accurate labels, but would be much slower.

We use a different prompt depending on whether timestep $t = 0$ or $t > 0$.
For $t = 0$, we simply provide the initial image-based state $s_0^{\text{img}}$ and a prompt (shown below) that asks the model to label all the values of atoms listed in the prompt with either ``True', ``False'' or ``Unknown''.

\begin{tcolorbox}[breakable, colback=white, colframe=black, boxrule=0.5pt, arc=0pt, left=5pt, right=5pt, top=5pt, bottom=5pt]
You are a vision system for a robot. Your job is to output the values of the following predicates based on the provided visual scene. For each predicate, output True, False, or Unknown if the relevant objects are not in the scene or the value of the predicate simply cannot be determined. Output each predicate value as a bulleted list with each predicate and value on a different line. For each output value, provide an explanation as to why you labelled this predicate as having this particular value. Use the format: $<$predicate$>$: $<$truth\_value$>$. $<$explanation$>$.

Predicates:
\end{tcolorbox}

We then specify the ground atom strings whose values we'd like to query. For instance, in one of the Burger tasks, some of these might be:

\begin{tcolorbox}[breakable, colback=white, colframe=black, boxrule=0.5pt, arc=0pt, left=5pt, right=5pt, top=5pt, bottom=5pt]
\noindent on\_table(patty1)\\
on\_table(patty2)\\
on\_table(top\_bun1)\\
on\_table(top\_bun2)\\
prepared(patty1)\\
prepared(patty2)\\
raw(patty1)\\
raw(patty2)\\
uncooked(patty1)\\
uncooked(patty2)
\end{tcolorbox}

\begin{figure*}[!ht]
\includegraphics[width=\textwidth]{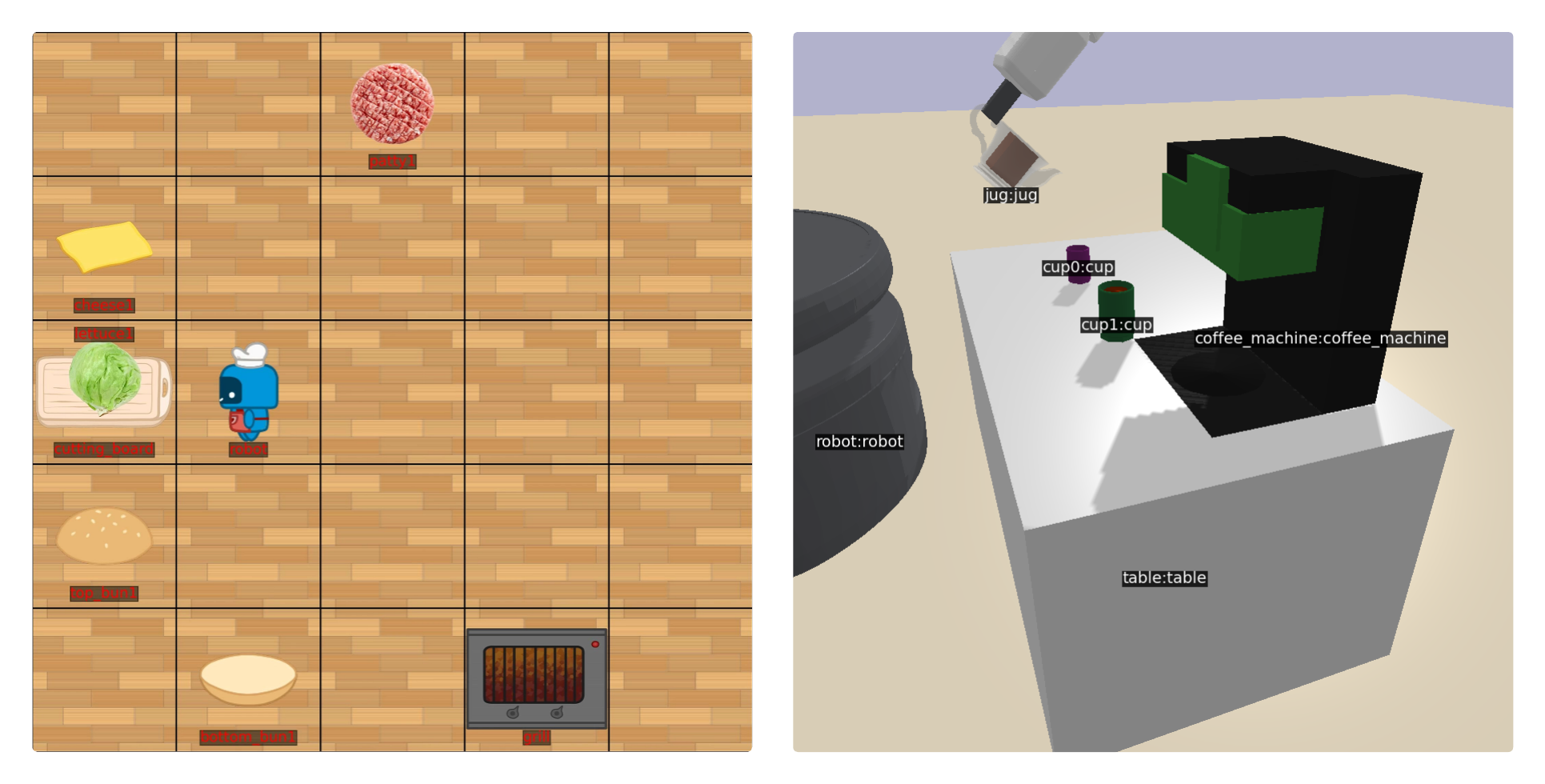}
\centering
\caption{\small{\textbf{Example object annotations in Burger and Coffee}. 
}}
\label{fig:object_annotation}
\vspace{-3mm}
\end{figure*}

Two aspects of our prompting strategy proved essential for achieving high labeling accuracy.
First, chain-of-thought prompting~\citep{wei2022chain} was extremely useful: our prompt (for $t > 0$) asks the VLM to explicitly provide reasoning for its choices in addition to labeling the truth value. Prompting the VLM to describe differences between two scenes in a chain-of-thought fashion led to much more accurate output than simply labeling atoms based on a single scene.
Second, we found that a form of set-of-marks prompting~\citep{yang2023set} where we annotate objects in the image with their names to be critical to the VLM being able to correctly identify objects being referred to in the text prompt. We only perform this annotation in the simulation environments, as we found it to be unnecessary and additional engineering effort in the real-world environments. Thus, in addition to assuming we have been given a set of user-specified objects (with names, an optional type, and descriptor string), we also assume, for the simulation environments (in which all objects have a type), that we have access to an object detection system that can classify objects according to their type, and track individual object instances over time, disambiguating between multiple objects of the same type across frames. An example of such annotations in the Burger and Coffee environments is shown in Figure~\ref{fig:object_annotation}. Notice that we differentiate between "cup0" and "cup1". In the simulation environments, an annotation for an object is just the type of the object plus a number. 

For all states $s_t$ in a trajectory that are not the initial state (i.e., $t > 0$), we prompt the VLM with the following information: (1) the image-based state from the current and previous timesteps (i.e. $s_t^{\text{img}}$ and $s_{t-1}^{\text{img}}$), (2) the action executed to get to the current state (i.e. $a_{t-1})$, and (3) the response of the VLM from the previous timestep. Here is an example of a prompt in one of the Burger tasks, at particular timestep.

\begin{tcolorbox}[breakable, colback=white, colframe=black, boxrule=0.5pt, arc=0pt, left=5pt, right=5pt, top=5pt, bottom=5pt]
You are a vision system for a robot. You are provided with two images corresponding to the states before and after a particular skill is executed. You are given a list of predicates below, and you are given the values of these predicates in the image before the skill is executed. Your job is to output the values of the following predicates in the image after the skill is executed. Pay careful attention to the visual changes between the two images to figure out which predicates change and which predicates do not change. For the predicates that change, list these separately at the end of your response. Note that in some scenes, there might be no changes. First, output a description of what changes you expect to happen based on the skill that was just run, explicitly noting the skill that was run. Second, output a description of what visual changes you see happen between the before and after images, looking specifically at the objects involved in the skill's arguments, noting what objects these are. Next, output each predicate value in the after image as a bulleted list with each predicate and value on a different line. For each predicate value, provide an explanation as to why you labeled this predicate as having this particular value. Use the format: $<$predicate$>$: $<$truth\_value$>$. $<$explanation$>$.

Your response should have three sections. Here is an outline of what your response should look like:

[START OUTLINE]

\# Expected changes based on the executed skill

[insert your analysis on the expected changes you will see based on the skill that was executed]

\# Visual changes observed between the images

[insert your analysis on the visual changes observed between the images]

\# Predicate values in the after image

[insert your bulleted list of `* <predicate>: <truth value>. <explanation>`]

[END OUTLINE]

Predicates: \\
available(robot)\\
busy(robot)\\
cooked(patty1)\\
empty(bottom\_bun1)\\
empty(grill)\\
empty(robot)\\
empty\_grill(grill)\\
free(robot)\\

Skill executed between states: Place[robot:robot, top\_bun1:top\_bun, patty1:patty]

Predicate values in the first scene, before the skill was executed: 
\# Expected changes based on the executed skill

The skill executed is "Pick[robot:robot, top\_bun1:top\_bun]". We expect the robot to pick up the top bun. The top bun should no longer be on the ground or table and should be in the robot's possession.

\# Visual changes observed between the images

In the before image, the top bun is on the ground. In the after image, the robot is holding the top bun, indicating that the robot has successfully picked it up.

\# Predicate values in the after image

* available(robot): Unknown. The previous value was unknown, and the pick action does not provide information about availability.\\
* busy(robot): Unknown. The previous value was unknown, and the pick action does not provide information about busyness.\\
* cooked(patty1): True. The patty remains cooked, as it was cooked before the action. Previously, the value was true.\\
* empty(bottom\_bun1): False. The bottom bun remains unchanged. Previously, the value was false.\\
* empty(grill): True. The grill remains empty, as the action does not involve the grill. Previously, the value was true.\\
* empty(robot): False. The robot is now holding the top bun. Previously, the value was true.\\
* empty\_grill(grill): True. The grill remains empty, as the action does not involve the grill. Previously, the value was true.\\
* free(robot): Unknown. The previous value was unknown, and the pick action does not provide information about freedom.
...
\end{tcolorbox}

There are two additional prompting techniques we found helpful and important for labeling accuracy: (1) asking the VLM to ``double-check'' its label output based on the explanations it provided, and (2) augmenting the image-based state to include a close-up crop of the image for the objects the robot is currently interacting with. We perform (2) only in the Burger domain.
For double-checking, we provide the VLM with the following prompt:

\begin{tcolorbox}[breakable, colback=white, colframe=black, boxrule=0.5pt, arc=0pt, left=5pt, right=5pt, top=5pt, bottom=5pt]
Sometimes your reasoning about the value of a predicate at the current timestep uses an incorrect value of that predicate in the previous timestep. Below, I give you give you the values of the predicates at the previous timestep once again. Please check your reasoning and provide a corrected version of your previous answer, if it needs correcting. Regardless of whether or not it needs correctly, your reply should be formatted exactly the same as the previous answer.
\end{tcolorbox}

We use double-checking only at training time to maximize labeling accuracy before optimization.

\subsubsection{Atom Proposal} 
Recall from Section~\ref{subsec:vlm-predicates} that the objective of atom proposal is to generate an initial pool of visual predicates given a set of demonstrations $\D$. 
We prompt a VLM to propose ground atoms on each demonstration $d \in D$, and then aggregate these and ``lift'' them into predicates.
Importantly, it is crucial to have diversity in the initial predicate pool: our approach will subselect and remove irrelevant predicates, but will fail if the initial pool does not have a predicate that is important for decision-making across tasks from the task distribution $\T$.
To promote diversity in the pool, we explicitly prompt the VLM to list all ground atoms that might be relevant to decision-making, and also to output at least one synonym and antonym of every ground atom it proposes (e.g. if it proposes `cooked(patty1)', it might also propose `grilled(patty1)' and `uncooked(patty1)'). 

Below, we provide our full prompt for atom proposal:
\begin{tcolorbox}[breakable, colback=white, colframe=black, boxrule=0.5pt, arc=0pt, left=5pt, right=5pt, top=5pt, bottom=5pt]
You are a robotic vision system whose job is to output a structured set of predicates useful for describing important concepts in the following demonstration of a task. You will be provided with a list of actions used during the task, as well as images of states before and after every action execution. Please provide predicates in terms of the following objects: \{objs\}. For each predicate, output it in the following format: predicate\_name(obj1, obj2, obj3...). Start by generating predicates that change before and after each action. After this, generate any other predicates that perhaps do not change but are still important to describing the demonstration shown. For each predicate you generate, also generate some predicates that are synonyms and antonyms so that any predicate that is even tangentially relevant to the demonstrations is generated.
\end{tcolorbox}
We then provide a text representation of the action sequence executed in the demonstration trajectory.
An example in a Burger task is:
\begin{tcolorbox}[breakable, colback=white, colframe=black, boxrule=0.5pt, arc=0pt, left=5pt, right=5pt, top=5pt, bottom=5pt]
\noindent Skills executed in trajectory:\\
Pick[robot:robot, patty1:patty]\\
Place[robot:robot, patty1:patty, grill:grill]\\
Cook[robot:robot, patty1:patty, grill:grill]\\
Pick[robot:robot, patty1:patty]\\
Place[robot:robot, patty1:patty, bottom\_bun1:bottom\_bun]\\
Pick[robot:robot, top\_bun1:top\_bun]\\
Place[robot:robot, top\_bun1:top\_bun, patty1:patty]
\end{tcolorbox}

An example response from the VLM in a Burger task is shown below. 

\begin{tcolorbox}[breakable, colback=white, colframe=black, boxrule=0.5pt, arc=0pt, left=5pt, right=5pt, top=5pt, bottom=5pt]

\textbf{Predicates for Each Action}

\begin{enumerate}
    \item \textbf{Pick[robot:robot, patty1:patty]}
    \begin{itemize}
        \item Before: on(patty1, cutting\_board)
        \item After: holding(robot, patty1)
        \item Synonyms: grasping(robot, patty1), carrying(robot, patty1)
        \item Antonyms: releasing(robot, patty1), dropping(robot, patty1)
    \end{itemize}

    \item \textbf{Place[robot:robot, patty1:patty, grill:grill]}
    \begin{itemize}
        \item Before: holding(robot, patty1)
        \item After: on(patty1, grill)
        \item Synonyms: atop(patty1, grill), above(patty1, grill)
        \item Antonyms: off(patty1, grill), under(patty1, grill)
    \end{itemize}

    \item \textbf{Cook[robot:robot, patty1:patty, grill:grill]}
    \begin{itemize}
        \item Before: raw(patty1)
        \item After: cooked(patty1)
        \item Synonyms: grilled(patty1), prepared(patty1)
        \item Antonyms: raw(patty1), uncooked(patty1)
    \end{itemize}

    \item \textbf{Pick[robot:robot, patty1:patty]}
    \begin{itemize}
        \item Before: on(patty1, grill)
        \item After: holding(robot, patty1)
        \item Synonyms: grasping(robot, patty1), carrying(robot, patty1)
        \item Antonyms: releasing(robot, patty1), dropping(robot, patty1)
    \end{itemize}

    \item \textbf{Place[robot:robot, patty1:patty, bottom\_bun1:bottom\_bun]}
    \begin{itemize}
        \item Before: holding(robot, patty1)
        \item After: on(patty1, bottom\_bun1)
        \item Synonyms: atop(patty1, bottom\_bun1), above(patty1, bottom\_bun1)
        \item Antonyms: off(patty1, bottom\_bun1), under(patty1, bottom\_bun1)
    \end{itemize}

    \item \textbf{Pick[robot:robot, top\_bun1:top\_bun]}
    \begin{itemize}
        \item Before: on(top\_bun1, cutting\_board)
        \item After: holding(robot, top\_bun1)
        \item Synonyms: grasping(robot, top\_bun1), carrying(robot, top\_bun1)
        \item Antonyms: releasing(robot, top\_bun1), dropping(robot, top\_bun1)
    \end{itemize}

    \item \textbf{Place[robot:robot, top\_bun1:top\_bun, patty1:patty]}
    \begin{itemize}
        \item Before: holding(robot, top\_bun1)
        \item After: on(top\_bun1, patty1)
        \item Synonyms: atop(top\_bun1, patty1), above(top\_bun1, patty1)
        \item Antonyms: off(top\_bun1, patty1), under(top\_bun1, patty1)
    \end{itemize}
\end{enumerate}

\vspace{1em}\textbf{Other Important Predicates}

\textbf{Additional Initial State Predicates:}
\begin{itemize}
    \item on(bottom\_bun1, cutting\_board)
    \item on(cheese1, cutting\_board)
    \item on(lettuce1, cutting\_board)
    \item near(robot, grill)
    \item near(robot, bottom\_bun1)
    \item near(robot, top\_bun1)
\end{itemize}

\textbf{Synonyms:}
\begin{itemize}
    \item on(x, y): atop(x, y), above(x, y)
    \item holding(x, y): grasping(x, y), carrying(x, y)
    \item cooked(x): grilled(x), prepared(x)
    \item raw(x): uncooked(x), fresh(x)
    \item near(x, y): close\_to(x, y), adjacent\_to(x, y)
\end{itemize}

\textbf{Antonyms:}
\begin{itemize}
    \item on(x, y): off(x, y), under(x, y)
    \item holding(x, y): releasing(x, y), dropping(x, y)
    \item cooked(x): raw(x), uncooked(x)
    \item raw(x): cooked(x), prepared(x)
    \item near(x, y): far\_from(x, y), distant\_from(x, y)
\end{itemize}

\end{tcolorbox}

We do the same set-of-marks prompting that we do for atom labeling, for atom proposal. Also, note that in the Coffee environment, we add "Generate only 5-10 predicates" to the proposal prompt to help restrict the total number of predicates the VLM proposes. This alleviates some of the difficulty the hill-climbing procedure would normally have to deal with, when faced with a very large set of predicates (e.g. 200+). Note that the VLM proposes atoms for each trajectory, so when we ask it to generate 5-10 atoms, it generates between $5n$ and $10n$ atoms in total if we had n demonstrations.

\subsection{Additional Experimental Details}
\label{appendix:additional-experiments}
\subsubsection{Experimental Evaluation and Hyperparameters.} 
We used the following hyperparameter values in our different domains. Where applicable, these values were determined via a validation set.
Here $N_\text{demo}$ is the size of the training dataset.

\begin{table}[h]
\centering
\begin{tabular}{lcccc}
\toprule
\textbf{Task} & \textbf{$J_{\text{thresh}}$} & \textbf{$h_{\text{pre\_frac}}$} & \textbf{$h_{\text{data\_frac}}$} & \textbf{$N_{\text{demo}}$} \\
\midrule
Kitchen               & 100  & 0.8 & 0.05 & 3  \\
Bigger Burger         & 2000 & 0.8 & 0.05 & 12 \\
More Burger Stacks    & 2000 & 0.8 & 0.05 & 12 \\
Combo Burger          & 2000 & 0.8 & 0.05 & 12 \\
Coffee                & 100   & 0.8 & 0.05 & 5  \\
Cleanup               & 100  & 1.0 & 0.4  & 6  \\
Juice                 & 0    & 1.0 & 0.4 & 10 \\
\bottomrule
\end{tabular}
\vspace{1em}
\caption{Domain parameters used in experiments.}
\label{tab:task-params}
\end{table}

\subsubsection{Additional Environment Details.} Here, we describe in detail the initial predicates, training demonstrations, and test tasks for each task in each of our experimental environments. Note that the controllers listed below have discrete object parameters (indicated by the `\texttt{?}') as well as continuous parameters where applicable $\theta$ (shown within []).

\begin{figure}[!t]
\includegraphics[width=\textwidth]{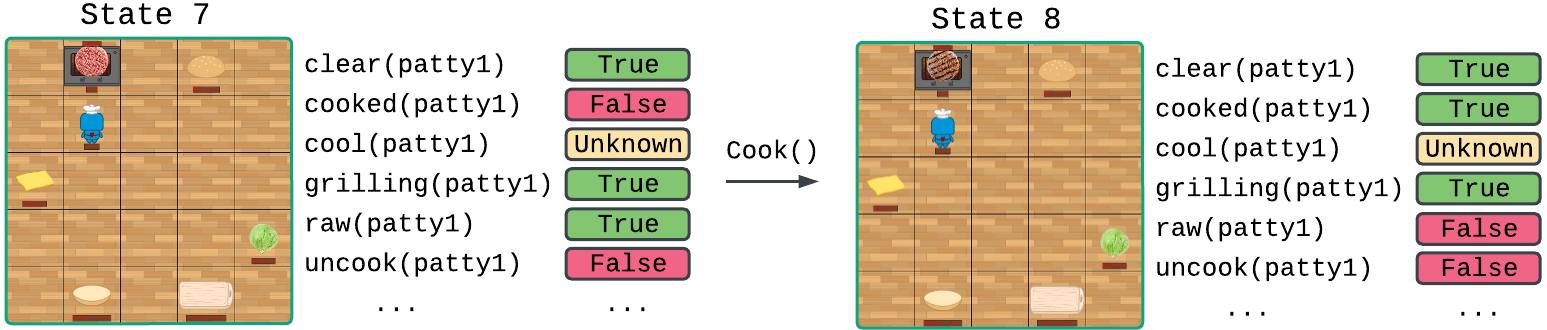}
\centering
\caption{\small{\textbf{Labeling of ground atoms with visual predicates involving the patty object type in Burger}. This figure depicts the visual state before and after the \textit{cook} skill in demonstration \#0 for the task distribution ``Combo Burger". It also shows the labels assigned by the VLM to ground atoms with visual predicates involving the patty object in these states. 
The VLM proposes a wide variety of predicates involving the patty: \texttt{clear}, \texttt{cooked}, \texttt{cool}, \texttt{grilling}, \texttt{raw}, and \texttt{uncook}. 
Of these, \texttt{cooked(?p:patty)} is most relevant to achieving the goal while also being accurately labeled. Our approach automatically selects this predicate from the pool. See Figure~\ref{fig:chop_labeling} in the appendix for an analogous situation with the \textit{chop} skill. 
}}
\label{fig:cooklabeling}
\vspace{-3mm}
\end{figure}
\begin{figure*}[!ht]
\includegraphics[width=\textwidth]{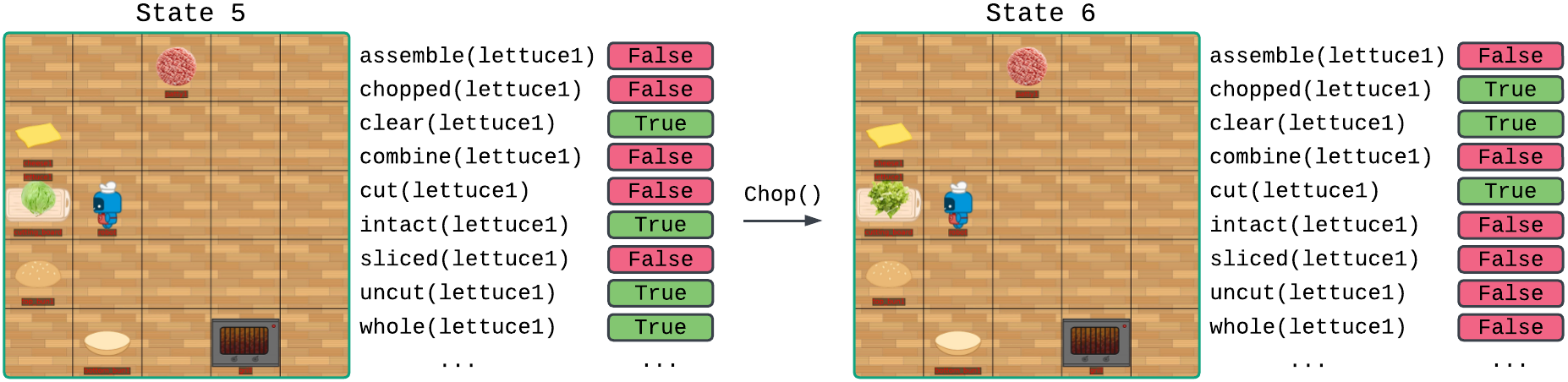}
\centering
\caption{\small{\textbf{Labeling of ground atoms with visual predicates involving the lettuce object type in Burger}. 
This figure depicts the visual state before and after the \textit{chop} skill in demonstration \#1 for the task distribution ``Combo Burger". 
It also shows the labels assigned by the VLM to ground atoms with visual predicates involving the lettuce object in these states. 
The VLM proposes a wide variety of predicates involving the patty: \texttt{assemble}, \texttt{chopped}, \texttt{clear}, \texttt{combine}, \texttt{cut}, \texttt{intact}, \texttt{sliced}, \texttt{uncut}, and \texttt{whole}.
Of these, \texttt{chopped(?l:lettuce)} and \texttt{cut(?l:lettuce)} are the  most relevant to achieving the goal while also being accurately labeled. Our approach automatically selects the former predicate from the pool.
}}
\label{fig:chop_labeling}
\vspace{-3mm}
\end{figure*}

\begin{tightlist}
    \item \textit{Kitchen:}
    \begin{tightlist}
        \item Initial predicates: 
        \begin{tightlist}
            \item \texttt{KettleBoiling(?k:kettle, ?kn:knob, ?b:burner)}: is true only if ?kn is turned on, and the kettle ?k is ontop of ?b, and if ?kn and ?b are 'linked' (i.e., ?kn is the knob that causes burner ?b to glow red with heat).
        \end{tightlist}
        \item Skills:
        \begin{tightlist}
            \item \texttt{TurnOnKnob(?g:gripper, ?kn:knob, [push\_dir])}: Moves the gripper ?g to a fixed location near the knob, and pushes with angle ``push\_dir'' with respect to the horizontal axis of the gripper for a fixed number of timesteps in an effort to flick the corresponding knob on.
            \item \texttt{PushKettleOntoBurner(?g:gripper, ?k:kettle, ?b:burner, [push\_x. push\_y, push\_z])}: Moves the gripper behind the current location of kettle ?k and then pushes along the 3D vector [push\_x. push\_y, push\_z] for a fixed number of timesteps.
        \end{tightlist}
        \item Training demonstrations: we provide 3 demonstrations that execute \texttt{TurnOnKnob} and then \texttt{PushKettleOntoBurner} in sequence to achieve  \texttt{KettleBoiling(kettle1, knob2, burner2)}, where burner2 and knob2 are in the back left on the stove. The kettle starts out on the front left of the stove.
        \item Test tasks: Given the kettle starts out on the front right burner, we task the agent with moving it to the back right burner. In this case, plans are only two steps, and actually simply replaying the demonstration plans will work.
    \end{tightlist}
    \item \textit{Burger:} The three different task distributions we implement in this domain share common skills. We list these before listing the task-distribution-specific initial predicates, demonstrations, and goals. 

    General info:
    \begin{tightlist}
        \item A ``burger" consists of a top bun above a bottom bun, with one or more items in between.
        \item Every type has a row/column/z attribute. The z attribute changes when an object is picked up or placed onto other objects. The grill and cutting board are of type \texttt{object}; the patty, lettuce, cheese, bottom bun, and top bun are of type \texttt{item}, which is a subtype of type \texttt{object}; the robot is of type \texttt{robot}, and has additional attributes ``fingers" that indicates how open its gripper is, and ``dir" that indicates the direction it is facing.
    \end{tightlist}

    Skills:
    \begin{tightlist}
        \item \texttt{Pick(?r:robot, ?i:item, [])}. Moves the robot \texttt{?r} to a cell adjacent to item \texttt{?i} and picks it up if the robot isn't currently holding anything.
        \item \texttt{Place(?r:robot, ?i:item, ?o:object, [])}. Moves the robot \texttt{?r} to a cell adjacent to object \texttt{?o} and places item \texttt{?i} atop object \texttt{?o} if it is holding item \texttt{?i}.
        \item \texttt{Cook(?r:robot, ?p:patty, ?g:grill, [])}. Given the robot is adjacent to grill \texttt{?g} and patty \texttt{?p} is atop the grill, cooks the patty so it appears grilled.
        \item \texttt{Chop(?r:robot, ?l:lettuce, ?c:cutting\_board, [])}. Given the robot is adjacent to cutting board \texttt{?c} and lettuce \texttt{?l} is atop the cutting board, chops the lettuce with a knife that's on the cutting board so it appears chopped.
    \end{tightlist}

    \begin{tightlist}
        \item \textbf{Bigger Burger}
        \begin{tightlist}
            \item Initial predicates: 
            \begin{tightlist}
                \item \texttt{On(?o1:object, ?o2:object)}: turns true when \texttt{?o1} is atop \texttt{?o2}.
                \item \texttt{OnGround(?o1:object)}: turns true when \texttt{?o1} is atop a cell \texttt{?o2} that forms the ground.
                \item \texttt{Clear(?o:object)}: is true only when there is no object atop the object \texttt{?o}.
                \item \texttt{Holding(?r:robot, ?i:item)}: turns true only when the robot \texttt{?r} is holding the item \texttt{?i}.
                \item \texttt{SomewhereAboveAndPrepped(?p:patty, ?b:bottom\_bun)}: turns true when \texttt{?p} is somewhere above \texttt{?b} and \texttt{?p} is cooked.
                \item \texttt{RightAboveAndPrepped(?p:patty, ?c:cutting\_board)}: turns true when \texttt{?p} is right above \texttt{?c} and \texttt{?p} is cooked.
                \item \texttt{RightAboveAndPrepped(?p:patty, ?g:grill)}: turns true when \texttt{?p} is right above \texttt{?g} and \texttt{?p} is cooked.
                \item \texttt{RightAboveAndPrepped(?p1:patty, ?p2:patty)}: turns true when \texttt{?p1} is right above \texttt{?p2} and \texttt{?p1} is cooked.
            \end{tightlist}
            \item Training demonstrations: We provide 4 demonstrations that make one burger with a single cooked patty, 4 demonstrations that cook two patties and stack them on the cutting board, and 4 demonstrations that cook two patties and stack them on the grill.
            \item Test tasks: There are 10 test tasks. In 5 of these test tasks, we ask the agent to make a burger with 2 cooked patties in it. In the other 5 test tasks, we ask the agent to make a burger with 2 cooked patties in it in addition to making a single open-face burger with a cooked patty, and the agent starts out holding a raw patty.
        \end{tightlist}
        \item \textbf{More Burger Stacks}
        \begin{tightlist}
            \item Initial predicates: 
            \begin{tightlist}
                \item \texttt{On(?o1:object, ?o2:object)}: turns true when \texttt{?o1} is atop \texttt{?o2}.
                \item \texttt{OnGround(?o1:object)}: turns true when \texttt{?o1} is atop a cell \texttt{?o2} that forms the ground.
                \item \texttt{Clear(?o:object)}: is true only when there is no object atop the object \texttt{?o}.
                \item \texttt{Holding(?r:robot, ?i:item)}: turns true only when the robot \texttt{?r} is holding the item \texttt{?i}.
                \item \texttt{SomewhereAboveAndPrepped(?p:patty, ?b:bottom\_bun)}: turns true when \texttt{?p} is somewhere above \texttt{?b} and \texttt{?p} is cooked.
            \end{tightlist}
            \item Training demonstrations: We provide 1 demonstration that makes two burgers, each with a single cooked patty, and 11 demonstrations that make one burger with a single cooked patty. 
            \item Test tasks: There are 10 test tasks. In 5 of these test tasks, we ask the agent to make 5 ``open-face" burgers -- burgers that consist of a cooked patty on a bottom bun. In the other 5 test tasks, we ask the agent to make 6 open-face burgers, and the agent starts out holding a raw patty.
        \end{tightlist}
        \item \textbf{Combo Burger}
        \begin{tightlist}
            \item Initial predicates: 
            \begin{tightlist}
                \item \texttt{On(?o1:object, ?o2:object)}: turns true when \texttt{?o1} is atop \texttt{?o2}.
                \item \texttt{OnGround(?o1:object)}: turns true when \texttt{?o1} is atop a cell \texttt{?o2} that forms the ground.
                \item \texttt{Clear(?o:object)}: is true only when there is no object atop the object \texttt{?o}.
                \item \texttt{Holding(?r:robot, ?i:item)}: turns true only when the robot \texttt{?r} is holding the item \texttt{?i}.
                \item \texttt{SomewhereAboveAndPrepped(?p:patty, ?b:bottom\_bun)}: turns true when \texttt{?p} is somewhere above \texttt{?b} and \texttt{?p} is cooked.
                \item \texttt{SomewhereAboveAndPrepped(?l:lettuce, ?b:bottom\_bun)}: turns true when \texttt{?l} is somewhere above \texttt{?b} and \texttt{?l} is chopped.
                \item \texttt{SomewhereAboveAndPrepped(?l:lettuce, ?p:patty)}: turns true when \texttt{?l} is somewhere above \texttt{?p} and \texttt{?l} is chopped.
            \end{tightlist}
            \item Training demonstrations: We provide 3 demonstrations that make one burger with a single cooked patty, 3 demonstrations that make one burger with a single chopped lettuce, 3 demonstrations that place a raw patty on the cutting board, chop lettuce, and place the chopped lettuce on the patty, and 3 demonstrations that place a raw patty on the grill, chop lettuce, and place the chopped lettuce on the patty.
            \item Test tasks: There are 10 test tasks. In 5 of these test tasks, we ask the agent to make two burgers, each with chopped lettuce on a cooked patty. In the other 5 test tasks, we ask the agent to make two burgers, each with chopped lettuce on a cooked patty, in addition to making another burger with a single cooked patty, and the agent starts out holding a raw patty.
        \end{tightlist}
    \end{tightlist}
 
    \item \textit{Coffee:}
    \begin{tightlist}
        \item Initial predicates: 
        \begin{tightlist}
            \item \texttt{CupFilled(?c:cup)}: turns true when cup is filled with liquid above a certain threshold.
            \item \texttt{Holding(?r:robot, ?j:jug)}: turns true when robot is holding the jug.
            \item \texttt{OnTable(?j:jug)}: turns true when jug is on the table.
            \item \texttt{HandEmpty(?r:robot)}: turns true when the robot is not holding anything.
            \item \texttt{RobotAboveCup(?r:robot, ?c:cup)}: turns true when the robot is close enough to the cup to pour liquid into the cup from a jug.
            \item \texttt{JugAboveCup(?j:jug, ?c:cup)}: turns true when the jug is close enough to the cup to pour liquid into the cup.
            \item \texttt{NotAboveCup(?r:robot, ?j:jug)}: turns true when the robot is not close enough to the cup to pour liquid into the cup from a jug.
            \item \texttt{PressingButton(?r:robot, ?m:coffee\_machine)}: turns true when the robot is pressing the power button of the coffee machine.
            \item \texttt{NotSameCup(?c1:cup, ?c2:cup)}: turns true when the two cups in question are not the same object.
            \item \texttt{JugPickable(?j:jug)}: turns true when the handle of the jug is accessible so that the robot can pick it up, usually after the skill \textit{RotateItemUntilHandleAccessible} is run if not true already.
        \end{tightlist}
        \item Skills:
        \begin{tightlist}
            \item \texttt{PickJug(?r:robot, ?j:jug, [])}: moves the robot gripper behind the jug's handle in the y-direction, moves down in the z-direction, and then moves forward in the y-direction before finally grasping the jug by the handle. 
            \item \texttt{PlaceJugInMachine(?r:robot, ?j:jug, ?m:coffee\_machine, [])}: picks up the jug slightly so as to avoid friction with the table, moves the jug to a position under the coffee machine's dispenser, and then places the jug.
            \item \texttt{TurnMachineOnAndFill(?r:robot, ?m:coffee\_machine, [])}: moves the robot up to be level with the power button of the coffee machine in the z-direction and then moves forward in the y-direction to press it, and then moves backwards. Terminates when the machine has finished dispensing coffee.
            \item \texttt{PourSomeLiquid(?r:robot, ?j:jug, ?c:cup, [])}: moves the robot next to the cup and pours from the jug until the cup is filled. 
            \item \texttt{RotateItemUntilHandleAccessible(?r:robot, ?j:jug, [])}: moves the robot to the jug, rotates the jug until its handle is as close to the robot as possible, and then breaks contact with the jug. 
        \end{tightlist}
        \item Training demonstrations: we provide 4 demonstrations where a single cup is on the table and the robot fills this single cup with coffee, and 1 demonstration where two cups are on the table and the robot fills both of these cups with coffee.
        \item Test tasks: We task the robot with filling three cups on the table with coffee.
    \end{tightlist}

    \item \textit{Cleanup:}
    \begin{tightlist}
        \item Initial predicates:
        \begin{tightlist}
            \item \texttt{Holding(?r:rrobot, ?m:movable)}: turns true when a robot \texttt{?r} is holding item \texttt{?m}. We provide this predicate directly because it is not possible to infer it from the image-based state due to occlusion.
            \item \texttt{HandEmpty(?r:robot)}: turns true when the robot isn't holding any objects. We provide this predicate directly because it is not possible to infer it from the image-based state (since no onboard camera can consistently see the hand).
            \item \texttt{InsideContainer(?m:movable, ?t:trash\_can)}: true when the object \texttt{?m} is inside the trash can \texttt{?t}.
            \item \texttt{WipedOfMarkerScribbles(?tab:table)}: true if the table \texttt{?tab} has no obvious markings atop its surface.
        \end{tightlist}
        \item Skills:
            \begin{tightlist}
                \item \texttt{PickFromTop(?g:robot, ?o:movable, ?s:immovable, [])}: Moves the robot ?g to grasp the top of movable object ?o, which is supported by immovable surface ?s, and lifts it.
                \item \texttt{PlaceInside(?g:robot, ?o:movable, ?s:immovable, [])}: Moves the robot ?g holding object ?o to position it inside immovable container ?s and releases it.
                \item \texttt{PickFromFloor(?g:robot, ?o:movable, [])}: Moves the robot ?g to grasp movable object ?o directly from the floor and lifts it.
                \item \texttt{WipeAndContinueHoldingEraser(?g:robot, ?o:movable, ?t:table, [])}: Uses the robot ?g holding movable object ?o (e.g., an eraser) to wipe the surface of table ?t while continuing to hold it.
                \item \texttt{DumpContentsOntoFloor(?g:robot, ?tc:trash\_can, [])}: Moves the robot ?g to tilt trash can ?tc and dump its contents onto the floor.
            \end{tightlist}
        \item Training demonstrations: We collect a dataset of 6 demonstrations. In 4 of these, the demonstrator clears an object from a table surface, then picks up an eraser and uses it to wipe a particular table of markings. In the remaining 2, the demonstrator dumps an object out of a narrow bin in order to pick it up and put it inside another bin.
        \item Testing tasks: As noted in Section~\ref{sec:experiments}, we test 5 different task variants. For the `New object instances' variant, we use a different table, with a different object on its surface and task the robot with achieving  \texttt{WipeAndContinueHoldingEraser} for the new table. For the `New visual background' variant, we use a different room. For `More objects', we task the robot with placing multiple objects into a particular bin. For `Novel goal 1', we setup an initial state where an object is on the floor (not atop the table), and task the robot with placing that object, as well as the eraser use for wiping, in the bin. For `Novel goal 2', we setup an initial state where the eraser itself is inside a narrow bin: the robot must first dump the bin to obtain the eraser (never seen at training time), and then use it to wipe the surface before then putting the eraser back in the bin. Please see our supplementary video for a recording of the robot performing this task.
    \end{tightlist}

    \item \textit{Juice:}
    \begin{tightlist}
        \item Initial predicates:
        \begin{tightlist}
            \item \texttt{Holding(?r:robot, ?m:movable)}: turns true when a robot \texttt{?r} is holding item \texttt{?m}. We provide this predicate directly because it is not possible to infer it from the image-based state due to occlusion.
            \item \texttt{HandEmpty(?r:robot)}: turns true when the robot isn't holding any objects. We provide this predicate directly because it is not possible to infer it from the image-based state (since no onboard camera can consistently see the hand).
            \item \texttt{JuiceInCup(?j:movable, ?c:container)}: true when the juice object \texttt{?j} is detected to be inside the container \texttt{?c}, as determined by the VLM.
            \item \texttt{Empty(?c:container)}: true when the container \texttt{?c} is detected to be empty, as determined by the vision-language model.
            \item \texttt{Inside(?m:movable, ?j:juicer)}: true when the movable object \texttt{?m} is detected to be inside the juicer \texttt{?j}, as determined by the VLM.
        \end{tightlist}
        \item Skills:
        \begin{tightlist}
            \item \texttt{PickContainer(?g:robot, ?c:container, [])}: Moves the robot ?g to grasp and pick up the container ?c.
            \item \texttt{PickMovable(?g:robot, ?m:movable, [])}: Moves the robot ?g to grasp and pick up the movable object ?m.
            \item \texttt{PlaceOnLeft(?g:robot, ?c:container, ?j:juicer, [])}: Moves the robot ?g holding container ?c to place it inside the waste valve region of juicer ?j.
            \item \texttt{PlaceOnRight(?g:robot, ?c:container, ?j:juicer, [])}: Moves the robot ?g holding container ?c to place it inside the juice valve region of juicer ?j.
            \item \texttt{PlaceInside(?g:robot, ?m:movable, ?j:juicer, [])}: Moves the robot ?g holding movable object ?m to place it inside juicer ?j.
            \item \texttt{DumpFromOneIntoOther(?g:robot, ?src:container, ?dest:container, [])}: Moves the robot ?g to dump the contents of source container ?src into destination container ?dest.
            \item \texttt{CloseLid(?g:robot, ?j:juicer, [])}: Moves the robot ?g to close the lid of juicer ?j.
            \item \texttt{RunMachine(?g:robot, ?j:juicer, ?c1:container, ?c2:container, [])}: Moves the robot ?g to turn on and run the juicer ?j, using containers ?c1 and ?c2 to collect outputs.
        \end{tightlist}
            
        \item Training demonstrations: We collect a dataset of 10 demonstrations. 
        Several demonstrations showcase putting a fruit (either an orange or an apple) into the juicer while cups are present in both the juice and waste valve regions. A few demonstrations show one cup absent, and that being placed under a corresponding valve before the juicer is run. In some of the demonstrations, one of the cups is full, and the demonstrator dumps the cup into an empty bowl or other container to empty it.
        A few demonstrations show dumping separately, and a few others simply show placing a fruit inside the juicer.        
        \item Testing tasks: As noted in Section~\ref{sec:experiments}, we test 5 different task variants. For the `New object instances' variant, we use a different table, with a different fruit to be juiced. For the `New visual background' variant, we use a different room. For `More objects', we task the robot with placing multiple objects into the juicer. For `Novel goal 1', we setup an initial state where there is no cup under the waste valve of the juicer, and no fruit inside the juicer. To solve this task, the robot must place a cup in the waste valve region, and then put the fruit in the juicer and run it. For `Novel goal 2', we setup an initial state where there are two cups in the correct regions (waste and juice valve), but one of them is full. Thus, the robot must first dump contents from that cup (never seen at training time) and place it back before running the juicer to make juice.
    \end{tightlist}
\end{tightlist}



\subsection{Real-World Experiment Details}
\label{appendix:robot-details}
\subsubsection{Real-Robot System Implementation}
\label{appendix:real-robot-system-details}
We build on the planning and perception systems implemented by~\citet{kumar2024practice}. 
In particular, we reuse their perception pipeline (that leverages a combination of DETIC~\citep{zhou2022detecting} and SAM~\citep{kirillov2023segment}) to construct our real-world object-centric state $S^{\text{obj}}$.
We also reuse their localization and mapping pipeline in order to implement movement, which is used implicitly in several of the skills we define for our real-world environments.
This pipeline requires the environment be outfitted with April tags (\url{https://april.eecs.umich.edu/software/apriltag}), as can be seen in our supplementary videos (the tags are thus only used for the movement skills, and nothing else).
Finally, we reuse several of their samplers (e.g. for navigation, and for grasping).

Similar to~\citet{kumar2024practice}, all skills are implemented programmatically via low-level calls to the Boston Dynamics Spot SDK (\url{https://dev.bostondynamics.com/}).

We construct the image-based state $S^{\text{img}}$ by directly using the $6$ on-board RGB cameras on the Boston Dynamics Spot\footnote{We slightly postprocess images by rotating them to offset for any camera rotation as described here: \url{https://dev.bostondynamics.com/python/examples/get_image/readme}}.

For more details, please see our open-source code release (which will be made public upon acceptance of this paper).

\subsubsection{Robot Domains Additional Details.}
\label{appendix:real-world-robot-learning-details}
We collected training data for both our real-world domains (Cleanup and Juice) by taking images from the point-of-view of a human performing the task (Figure~\ref{fig:cleanup-domain-overview}).
We include an object `hand:robot' in each demonstration to indicate that the demonstrator's hand is performing actions.
However, a key challenge with the Boston Dynamics Spot robot is that none of the onboard cameras can see the hand itself. This means that any invented visual predicates that have to do with the hand will not transfer to the robot.
Thus, during initial pool proposal (Section~\ref{subsec:vlm-proposal}), we explicitly exclude any visual atoms proposed that mention the hand.
The only predicates that involve the hand are thus \texttt{HandEmpty}, and \texttt{Holding} since these are provided as part of the initial pool.

In both Cleanup and Juice, we found that the VLM proposed a very large number (sometimes over $1000$) atoms, leading to an extremely slow hill-climbing optimization to subselect predicates.
We thus modified the atom proposal prompt to have the VLM restrict itself to proposing no more than $25$ atoms per demonstration.

In the Juice domain, we found that the VLM struggled with accurately recognizing whether cups were below specific valves regardless of how the predicate was named (e.g. `UnderWasteValve', `ReadyToReceiveWaste', `WithinWasteRegion').
We thus manually demarcated and labelled (via directly marking the table surface) the waste and juice regions next to the juicer. We found this was necessary for learning a reasonable world model, and for accurately planning and labelling states at test time.


\subsection{Examples of Learned Predicates and Learned Action Models per Environment}
\label{appendix:examples-from-learning}
Note that predicates that end in a number are a result of the VLM proposing multiple ground atoms with the same name but with different arguments. For example, the VLM might propose two ground atoms, "Clear(jug0)" and "Clear(table0)". Because each predicate must have a unique set of typed object arguments, we rename these ground atoms to "Clear0(jug0)" and "Clear1(table0)" during the process of parsing predicates from the VLM's response, and would end up with two predicates, "Clear0" and "Clear1".

Further note that if you see numbers skipped in the names of the action models, that is because low-data operators were pruned, as described in Appendix~\ref{appendix:op-learning}.

\begin{tcolorbox}[
    title={\textbf{Example learned predicates and learned action models in Kitchen}},
    colback=gray!10,
    colframe=black,
    sharp corners=southwest,
    boxrule=0.8pt,
    breakable
  ]
  \lstset{
    basicstyle=\ttfamily\scriptsize,
    breaklines=true
  }
  \begin{lstlisting}
Learned predicates:
NOT-[[0:surface].z<=[idx_0]1.59]

Learned operators:
STRIPS-Op0:
  Parameters: [?x0:surface, ?x1:gripper, ?x2:knob]
  Preconditions: [KnobAndBurnerLinked(?x2:knob, ?x0:surface)]
  Add Effects: [NOT-[[0:surface].z<=[idx_0]1.59](?x0:surface)]
  Delete Effects: []
  Ignore Effects: []
  Skill: MoveAndTurnOnKnob(?x1:gripper, ?x2:knob)

STRIPS-Op1:
  Parameters: [?x0:surface, ?x1:gripper, ?x2:kettle, ?x3:knob]
  Preconditions: [KnobAndBurnerLinked(?x3:knob, ?x0:surface), NOT-[[0:surface].z<=[idx_0]1.59](?x0:surface)]
  Add Effects: [KettleBoiling(?x2:kettle, ?x0:surface, ?x3:knob)]
  Delete Effects: []
  Ignore Effects: []
  Skill: PushKettleOntoBurner(?x1:gripper, ?x2:kettle, ?x0:surface)
  \end{lstlisting}
\end{tcolorbox}

\begin{tcolorbox}[
    title={\textbf{Example learned predicates and learned action models in More Stacks Burger}},
    colback=gray!10,
    colframe=black,
    sharp corners=southwest,
    boxrule=0.8pt,
    breakable
  ]
  \lstset{
    basicstyle=\ttfamily\scriptsize,
    breaklines=true
  }
  \begin{lstlisting}
Learned Predicates:
cooked0(?p:patty)
empty_hands0(?r:robot)

Learned operators:
STRIPS-Op0:
  Parameters: [?x0:patty, ?x1:robot]
  Preconditions: [Clear(?x0:patty), Clear(?x1:robot), OnGround(?x0:patty), empty_hands0(?x1:robot)]
  Add Effects: [Holding(?x1:robot, ?x0:patty)]
  Delete Effects: [Clear(?x0:patty), OnGround(?x0:patty), empty_hands0(?x1:robot)]
  Ignore Effects: []
  Skill: Pick(?x1:robot, ?x0:patty)

STRIPS-Op1:
  Parameters: [?x0:grill, ?x1:patty, ?x2:robot]
  Preconditions: [Clear(?x0:grill), Clear(?x2:robot), Holding(?x2:robot, ?x1:patty)]
  Add Effects: [Clear(?x1:patty), On(?x1:patty, ?x0:grill), empty_hands0(?x2:robot)]
  Delete Effects: [Clear(?x0:grill), Holding(?x2:robot, ?x1:patty)]
  Ignore Effects: []
  Skill: Place(?x2:robot, ?x1:patty, ?x0:grill)

STRIPS-Op2:
  Parameters: [?x0:grill, ?x1:patty, ?x2:robot]
  Preconditions: [Clear(?x1:patty), Clear(?x2:robot), On(?x1:patty, ?x0:grill), empty_hands0(?x2:robot)]
  Add Effects: [cooked0(?x1:patty)]
  Delete Effects: []
  Ignore Effects: []
  Skill: Cook(?x2:robot, ?x1:patty, ?x0:grill)

STRIPS-Op3:
  Parameters: [?x0:grill, ?x1:patty, ?x2:robot]
  Preconditions: [Clear(?x1:patty), Clear(?x2:robot), On(?x1:patty, ?x0:grill), cooked0(?x1:patty), empty_hands0(?x2:robot)]
  Add Effects: [Clear(?x0:grill), Holding(?x2:robot, ?x1:patty)]
  Delete Effects: [Clear(?x1:patty), On(?x1:patty, ?x0:grill), empty_hands0(?x2:robot)]
  Ignore Effects: []
  Skill: Pick(?x2:robot, ?x1:patty)

STRIPS-Op4:
  Parameters: [?x0:bottom_bun, ?x1:patty, ?x2:robot]
  Preconditions: [Clear(?x0:bottom_bun), Clear(?x2:robot), Holding(?x2:robot, ?x1:patty), OnGround(?x0:bottom_bun), cooked0(?x1:patty)]
  Add Effects: [Clear(?x1:patty), SomewhereAboveAndPrepped(?x0:bottom_bun, ?x1:patty), On(?x1:patty, ?x0:bottom_bun), empty_hands0(?x2:robot)]
  Delete Effects: [Clear(?x0:bottom_bun), Holding(?x2:robot, ?x1:patty)]
  Ignore Effects: []
  Skill: Place(?x2:robot, ?x1:patty, ?x0:bottom_bun)

STRIPS-Op5:
  Parameters: [?x0:robot, ?x1:top_bun]
  Preconditions: [Clear(?x0:robot), Clear(?x1:top_bun), OnGround(?x1:top_bun), empty_hands0(?x0:robot)]
  Add Effects: [Holding(?x0:robot, ?x1:top_bun)]
  Delete Effects: [Clear(?x1:top_bun), OnGround(?x1:top_bun), empty_hands0(?x0:robot)]
  Ignore Effects: []
  Skill: Pick(?x0:robot, ?x1:top_bun)

STRIPS-Op6:
  Parameters: [?x0:patty, ?x1:robot, ?x2:top_bun]
  Preconditions: [Clear(?x0:patty), Clear(?x1:robot), Holding(?x1:robot, ?x2:top_bun), cooked0(?x0:patty)]
  Add Effects: [Clear(?x2:top_bun), On(?x2:top_bun, ?x0:patty), empty_hands0(?x1:robot)]
  Delete Effects: [Clear(?x0:patty), Holding(?x1:robot, ?x2:top_bun)]
  Ignore Effects: []
  Skill: Place(?x1:robot, ?x2:top_bun, ?x0:patty)
  \end{lstlisting}
\end{tcolorbox}

\begin{tcolorbox}[
    title={\textbf{Example learned predicates and learned action models in Bigger Burger}},
    colback=gray!10,
    colframe=black,
    sharp corners=southwest,
    boxrule=0.8pt,
    breakable
  ]
  \lstset{
    basicstyle=\ttfamily\scriptsize,
    breaklines=true
  }
  \begin{lstlisting}
Learned Predicates:
cooked0(?p:patty)
[[0:robot].fingers<=[idx_0]0.5]

Learned operators:
STRIPS-Op0:
  Parameters: [?x0:patty, ?x1:robot]
  Preconditions: [Clear(?x0:patty), Clear(?x1:robot), OnGround(?x0:patty), [[0:robot].fingers<=[idx_0]0.5](?x1:robot)]
  Add Effects: [Holding(?x1:robot, ?x0:patty)]
  Delete Effects: [Clear(?x0:patty), OnGround(?x0:patty), [[0:robot].fingers<=[idx_0]0.5](?x1:robot)]
  Ignore Effects: []
  Skill: Pick(?x1:robot, ?x0:patty)

STRIPS-Op1:
  Parameters: [?x0:grill, ?x1:patty, ?x2:robot]
  Preconditions: [Clear(?x0:grill), Clear(?x2:robot), Holding(?x2:robot, ?x1:patty)]
  Add Effects: [Clear(?x1:patty), On(?x1:patty, ?x0:grill), [[0:robot].fingers<=[idx_0]0.5](?x2:robot)]
  Delete Effects: [Clear(?x0:grill), Holding(?x2:robot, ?x1:patty)]
  Ignore Effects: []
  Skill: Place(?x2:robot, ?x1:patty, ?x0:grill)

STRIPS-Op2:
  Parameters: [?x0:grill, ?x1:patty, ?x2:robot]
  Preconditions: [Clear(?x1:patty), Clear(?x2:robot), On(?x1:patty, ?x0:grill), [[0:robot].fingers<=[idx_0]0.5](?x2:robot)]
  Add Effects: [RightAboveAndPrepped(?x0:grill, ?x1:patty), cooked0(?x1:patty)]
  Delete Effects: []
  Ignore Effects: []
  Skill: Cook(?x2:robot, ?x1:patty, ?x0:grill)

STRIPS-Op3:
  Parameters: [?x0:grill, ?x1:patty, ?x2:robot]
  Preconditions: [Clear(?x1:patty), Clear(?x2:robot), RightAboveAndPrepped(?x0:grill, ?x1:patty), On(?x1:patty, ?x0:grill), [[0:robot].fingers<=[idx_0]0.5](?x2:robot), cooked0(?x1:patty)]
  Add Effects: [Clear(?x0:grill), Holding(?x2:robot, ?x1:patty)]
  Delete Effects: [Clear(?x1:patty), RightAboveAndPrepped(?x0:grill, ?x1:patty), On(?x1:patty, ?x0:grill), [[0:robot].fingers<=[idx_0]0.5](?x2:robot)]
  Ignore Effects: []
  Skill: Pick(?x2:robot, ?x1:patty)

STRIPS-Op4:
  Parameters: [?x0:bottom_bun, ?x1:patty, ?x2:robot]
  Preconditions: [Clear(?x0:bottom_bun), Clear(?x2:robot), Holding(?x2:robot, ?x1:patty), OnGround(?x0:bottom_bun), cooked0(?x1:patty)]
  Add Effects: [Clear(?x1:patty), SomewhereAboveAndPrepped(?x0:bottom_bun, ?x1:patty), On(?x1:patty, ?x0:bottom_bun), [[0:robot].fingers<=[idx_0]0.5](?x2:robot)]
  Delete Effects: [Clear(?x0:bottom_bun), Holding(?x2:robot, ?x1:patty)]
  Ignore Effects: []
  Skill: Place(?x2:robot, ?x1:patty, ?x0:bottom_bun)

STRIPS-Op5:
  Parameters: [?x0:robot, ?x1:top_bun]
  Preconditions: [Clear(?x0:robot), Clear(?x1:top_bun), OnGround(?x1:top_bun), [[0:robot].fingers<=[idx_0]0.5](?x0:robot)]
  Add Effects: [Holding(?x0:robot, ?x1:top_bun)]
  Delete Effects: [Clear(?x1:top_bun), OnGround(?x1:top_bun), [[0:robot].fingers<=[idx_0]0.5](?x0:robot)]
  Ignore Effects: []
  Skill: Pick(?x0:robot, ?x1:top_bun)

STRIPS-Op6:
  Parameters: [?x0:patty, ?x1:robot, ?x2:top_bun]
  Preconditions: [Clear(?x0:patty), Clear(?x1:robot), Holding(?x1:robot, ?x2:top_bun), cooked0(?x0:patty)]
  Add Effects: [Clear(?x2:top_bun), On(?x2:top_bun, ?x0:patty), [[0:robot].fingers<=[idx_0]0.5](?x1:robot)]
  Delete Effects: [Clear(?x0:patty), Holding(?x1:robot, ?x2:top_bun)]
  Ignore Effects: []
  Skill: Place(?x1:robot, ?x2:top_bun, ?x0:patty)

STRIPS-Op7:
  Parameters: [?x0:cutting_board, ?x1:patty, ?x2:robot]
  Preconditions: [Clear(?x0:cutting_board), Clear(?x2:robot), Holding(?x2:robot, ?x1:patty), cooked0(?x1:patty)]
  Add Effects: [Clear(?x1:patty), RightAboveAndPrepped(?x0:cutting_board, ?x1:patty), On(?x1:patty, ?x0:cutting_board), [[0:robot].fingers<=[idx_0]0.5](?x2:robot)]
  Delete Effects: [Clear(?x0:cutting_board), Holding(?x2:robot, ?x1:patty)]
  Ignore Effects: []
  Skill: Place(?x2:robot, ?x1:patty, ?x0:cutting_board)

STRIPS-Op8:
  Parameters: [?x0:patty, ?x1:patty, ?x2:robot]
  Preconditions: [Clear(?x0:patty), Clear(?x2:robot), Holding(?x2:robot, ?x1:patty), cooked0(?x0:patty), cooked0(?x1:patty)]
  Add Effects: [Clear(?x1:patty), RightAboveAndPrepped(?x0:patty, ?x1:patty), On(?x1:patty, ?x0:patty), [[0:robot].fingers<=[idx_0]0.5](?x2:robot)]
  Delete Effects: [Clear(?x0:patty), Holding(?x2:robot, ?x1:patty)]
  Ignore Effects: []
  Skill: Place(?x2:robot, ?x1:patty, ?x0:patty)

STRIPS-Op9:
  Parameters: [?x0:cutting_board, ?x1:patty, ?x2:robot]
  Preconditions: [Clear(?x1:patty), Clear(?x2:robot), RightAboveAndPrepped(?x0:cutting_board, ?x1:patty), On(?x1:patty, ?x0:cutting_board), [[0:robot].fingers<=[idx_0]0.5](?x2:robot), cooked0(?x1:patty)]
  Add Effects: [Clear(?x0:cutting_board), Holding(?x2:robot, ?x1:patty)]
  Delete Effects: [Clear(?x1:patty), RightAboveAndPrepped(?x0:cutting_board, ?x1:patty), On(?x1:patty, ?x0:cutting_board), [[0:robot].fingers<=[idx_0]0.5](?x2:robot)]
  Ignore Effects: []
  Skill: Pick(?x2:robot, ?x1:patty)
  \end{lstlisting}
\end{tcolorbox}

\begin{tcolorbox}[
    title={\textbf{Example learned predicates and learned action models in Combo Burger}},
    colback=gray!10,
    colframe=black,
    sharp corners=southwest,
    boxrule=0.8pt,
    breakable
  ]
  \lstset{
    basicstyle=\ttfamily\scriptsize,
    breaklines=true
  }
  \begin{lstlisting}
Learned Predicates:
cooked0(?p:patty)
chopped0(?l:lettuce)
clear5(?r:robot)

Learned operators:
STRIPS-Op0:
  Parameters: [?x0:patty, ?x1:robot]
  Preconditions: [Clear(?x0:patty), Clear(?x1:robot), OnGround(?x0:patty), clear5(?x1:robot)]
  Add Effects: [Holding(?x1:robot, ?x0:patty)]
  Delete Effects: [Clear(?x0:patty), OnGround(?x0:patty), clear5(?x1:robot)]
  Ignore Effects: []
  Skill: Pick(?x1:robot, ?x0:patty)

STRIPS-Op1:
  Parameters: [?x0:grill, ?x1:patty, ?x2:robot]
  Preconditions: [Clear(?x0:grill), Clear(?x2:robot), Holding(?x2:robot, ?x1:patty)]
  Add Effects: [Clear(?x1:patty), On(?x1:patty, ?x0:grill), clear5(?x2:robot)]
  Delete Effects: [Clear(?x0:grill), Holding(?x2:robot, ?x1:patty)]
  Ignore Effects: []
  Skill: Place(?x2:robot, ?x1:patty, ?x0:grill)

STRIPS-Op2:
  Parameters: [?x0:grill, ?x1:patty, ?x2:robot]
  Preconditions: [Clear(?x1:patty), Clear(?x2:robot), On(?x1:patty, ?x0:grill), clear5(?x2:robot)]
  Add Effects: [cooked0(?x1:patty)]
  Delete Effects: []
  Ignore Effects: []
  Skill: Cook(?x2:robot, ?x1:patty, ?x0:grill)

STRIPS-Op3:
  Parameters: [?x0:grill, ?x1:patty, ?x2:robot]
  Preconditions: [Clear(?x1:patty), Clear(?x2:robot), On(?x1:patty, ?x0:grill), clear5(?x2:robot), cooked0(?x1:patty)]
  Add Effects: [Clear(?x0:grill), Holding(?x2:robot, ?x1:patty)]
  Delete Effects: [Clear(?x1:patty), On(?x1:patty, ?x0:grill), clear5(?x2:robot)]
  Ignore Effects: []
  Skill: Pick(?x2:robot, ?x1:patty)

STRIPS-Op4:
  Parameters: [?x0:bottom_bun, ?x1:patty, ?x2:robot]
  Preconditions: [Clear(?x0:bottom_bun), Clear(?x2:robot), Holding(?x2:robot, ?x1:patty), OnGround(?x0:bottom_bun), cooked0(?x1:patty)]
  Add Effects: [Clear(?x1:patty), SomewhereAboveAndPrepped(?x0:bottom_bun, ?x1:patty), On(?x1:patty, ?x0:bottom_bun), clear5(?x2:robot)]
  Delete Effects: [Clear(?x0:bottom_bun), Holding(?x2:robot, ?x1:patty)]
  Ignore Effects: []
  Skill: Place(?x2:robot, ?x1:patty, ?x0:bottom_bun)

STRIPS-Op5:
  Parameters: [?x0:robot, ?x1:top_bun]
  Preconditions: [Clear(?x0:robot), Clear(?x1:top_bun), OnGround(?x1:top_bun), clear5(?x0:robot)]
  Add Effects: [Holding(?x0:robot, ?x1:top_bun)]
  Delete Effects: [Clear(?x1:top_bun), OnGround(?x1:top_bun), clear5(?x0:robot)]
  Ignore Effects: []
  Skill: Pick(?x0:robot, ?x1:top_bun)

STRIPS-Op6:
  Parameters: [?x0:patty, ?x1:robot, ?x2:top_bun]
  Preconditions: [Clear(?x0:patty), Clear(?x1:robot), Holding(?x1:robot, ?x2:top_bun), cooked0(?x0:patty)]
  Add Effects: [Clear(?x2:top_bun), On(?x2:top_bun, ?x0:patty), clear5(?x1:robot)]
  Delete Effects: [Clear(?x0:patty), Holding(?x1:robot, ?x2:top_bun)]
  Ignore Effects: []
  Skill: Place(?x1:robot, ?x2:top_bun, ?x0:patty)

STRIPS-Op7:
  Parameters: [?x0:lettuce, ?x1:robot]
  Preconditions: [Clear(?x0:lettuce), Clear(?x1:robot), OnGround(?x0:lettuce), clear5(?x1:robot)]
  Add Effects: [Holding(?x1:robot, ?x0:lettuce)]
  Delete Effects: [Clear(?x0:lettuce), OnGround(?x0:lettuce), clear5(?x1:robot)]
  Ignore Effects: []
  Skill: Pick(?x1:robot, ?x0:lettuce)

STRIPS-Op8:
  Parameters: [?x0:cutting_board, ?x1:lettuce, ?x2:robot]
  Preconditions: [Clear(?x0:cutting_board), Clear(?x2:robot), Holding(?x2:robot, ?x1:lettuce)]
  Add Effects: [Clear(?x1:lettuce), On(?x1:lettuce, ?x0:cutting_board), clear5(?x2:robot)]
  Delete Effects: [Clear(?x0:cutting_board), Holding(?x2:robot, ?x1:lettuce)]
  Ignore Effects: []
  Skill: Place(?x2:robot, ?x1:lettuce, ?x0:cutting_board)

STRIPS-Op9:
  Parameters: [?x0:cutting_board, ?x1:lettuce, ?x2:robot]
  Preconditions: [Clear(?x1:lettuce), Clear(?x2:robot), On(?x1:lettuce, ?x0:cutting_board), clear5(?x2:robot)]
  Add Effects: [chopped0(?x1:lettuce)]
  Delete Effects: []
  Ignore Effects: []
  Skill: Chop(?x2:robot, ?x1:lettuce, ?x0:cutting_board)

STRIPS-Op10:
  Parameters: [?x0:cutting_board, ?x1:lettuce, ?x2:robot]
  Preconditions: [Clear(?x1:lettuce), Clear(?x2:robot), On(?x1:lettuce, ?x0:cutting_board), chopped0(?x1:lettuce), clear5(?x2:robot)]
  Add Effects: [Clear(?x0:cutting_board), Holding(?x2:robot, ?x1:lettuce)]
  Delete Effects: [Clear(?x1:lettuce), On(?x1:lettuce, ?x0:cutting_board), clear5(?x2:robot)]
  Ignore Effects: []
  Skill: Pick(?x2:robot, ?x1:lettuce)

STRIPS-Op11:
  Parameters: [?x0:bottom_bun, ?x1:lettuce, ?x2:robot]
  Preconditions: [Clear(?x0:bottom_bun), Clear(?x2:robot), Holding(?x2:robot, ?x1:lettuce), OnGround(?x0:bottom_bun), chopped0(?x1:lettuce)]
  Add Effects: [Clear(?x1:lettuce), SomewhereAboveAndPrepped(?x0:bottom_bun, ?x1:lettuce), On(?x1:lettuce, ?x0:bottom_bun), clear5(?x2:robot)]
  Delete Effects: [Clear(?x0:bottom_bun), Holding(?x2:robot, ?x1:lettuce)]
  Ignore Effects: []
  Skill: Place(?x2:robot, ?x1:lettuce, ?x0:bottom_bun)

STRIPS-Op12:
  Parameters: [?x0:lettuce, ?x1:robot, ?x2:top_bun]
  Preconditions: [Clear(?x0:lettuce), Clear(?x1:robot), Holding(?x1:robot, ?x2:top_bun), chopped0(?x0:lettuce)]
  Add Effects: [Clear(?x2:top_bun), On(?x2:top_bun, ?x0:lettuce), clear5(?x1:robot)]
  Delete Effects: [Clear(?x0:lettuce), Holding(?x1:robot, ?x2:top_bun)]
  Ignore Effects: []
  Skill: Place(?x1:robot, ?x2:top_bun, ?x0:lettuce)

STRIPS-Op13:
  Parameters: [?x0:grill, ?x1:lettuce, ?x2:robot]
  Preconditions: [Clear(?x0:grill), Clear(?x2:robot), Holding(?x2:robot, ?x1:lettuce), chopped0(?x1:lettuce)]
  Add Effects: [Clear(?x1:lettuce), On(?x1:lettuce, ?x0:grill), clear5(?x2:robot)]
  Delete Effects: [Clear(?x0:grill), Holding(?x2:robot, ?x1:lettuce)]
  Ignore Effects: []
  Skill: Place(?x2:robot, ?x1:lettuce, ?x0:grill)

STRIPS-Op14:
  Parameters: [?x0:cutting_board, ?x1:patty, ?x2:robot]
  Preconditions: [Clear(?x0:cutting_board), Clear(?x2:robot), Holding(?x2:robot, ?x1:patty)]
  Add Effects: [Clear(?x1:patty), On(?x1:patty, ?x0:cutting_board), clear5(?x2:robot)]
  Delete Effects: [Clear(?x0:cutting_board), Holding(?x2:robot, ?x1:patty)]
  Ignore Effects: []
  Skill: Place(?x2:robot, ?x1:patty, ?x0:cutting_board)

STRIPS-Op15:
  Parameters: [?x0:grill, ?x1:lettuce, ?x2:robot]
  Preconditions: [Clear(?x1:lettuce), Clear(?x2:robot), On(?x1:lettuce, ?x0:grill), chopped0(?x1:lettuce), clear5(?x2:robot)]
  Add Effects: [Clear(?x0:grill), Holding(?x2:robot, ?x1:lettuce)]
  Delete Effects: [Clear(?x1:lettuce), On(?x1:lettuce, ?x0:grill), clear5(?x2:robot)]
  Ignore Effects: []
  Skill: Pick(?x2:robot, ?x1:lettuce)

STRIPS-Op16:
  Parameters: [?x0:lettuce, ?x1:patty, ?x2:robot]
  Preconditions: [Clear(?x1:patty), Clear(?x2:robot), Holding(?x2:robot, ?x0:lettuce), chopped0(?x0:lettuce)]
  Add Effects: [Clear(?x0:lettuce), SomewhereAboveAndPrepped(?x1:patty, ?x0:lettuce), On(?x0:lettuce, ?x1:patty), clear5(?x2:robot)]
  Delete Effects: [Clear(?x1:patty), Holding(?x2:robot, ?x0:lettuce)]
  Ignore Effects: []
  Skill: Place(?x2:robot, ?x0:lettuce, ?x1:patty)

\end{lstlisting}
\end{tcolorbox}

\begin{tcolorbox}[
    title={\textbf{Example learned predicates and learned action models in Coffee}},
    colback=gray!10,  
    colframe=black, 
    sharp corners=southwest, 
    boxrule=0.8pt, 
    breakable
  ]
  \lstset{
    basicstyle=\ttfamily\scriptsize,
    breaklines=true
  }
  \begin{lstlisting}
Learned predicates:
CupReceivingLiquid0(?c:cup)
JugFilled0(?j:jug)
JugInMachine0(?j:jug, ?m:coffee_machine)

Learned operators:
STRIPS-Op0:
  Parameters: [?x0:cup, ?x1:cup, ?x2:jug, ?x3:robot]
  Preconditions: [HandEmpty(?x3:robot), JugPickable(?x2:jug), NotAboveCup(?x3:robot, ?x2:jug), NotSameCup(?x0:cup, ?x1:cup), NotSameCup(?x1:cup, ?x0:cup), OnTable(?x2:jug)]
  Add Effects: [Holding(?x3:robot, ?x2:jug), JugAboveCup(?x2:jug, ?x0:cup), JugAboveCup(?x2:jug, ?x1:cup), RobotAboveCup(?x3:robot, ?x0:cup), RobotAboveCup(?x3:robot, ?x1:cup)]
  Delete Effects: [HandEmpty(?x3:robot), NotAboveCup(?x3:robot, ?x2:jug), OnTable(?x2:jug)]
  Ignore Effects: []
  Skill: PickJug(?x3:robot, ?x2:jug)
STRIPS-Op1:
  Parameters: [?x0:coffee_machine, ?x1:cup, ?x2:cup, ?x3:jug, ?x4:robot]
  Preconditions: [Holding(?x4:robot, ?x3:jug), JugAboveCup(?x3:jug, ?x1:cup), JugAboveCup(?x3:jug, ?x2:cup), JugPickable(?x3:jug), NotSameCup(?x1:cup, ?x2:cup), NotSameCup(?x2:cup, ?x1:cup), RobotAboveCup(?x4:robot, ?x1:cup), RobotAboveCup(?x4:robot, ?x2:cup)]
  Add Effects: [HandEmpty(?x4:robot), JugInMachine0(?x3:jug, ?x0:coffee_machine), NotAboveCup(?x4:robot, ?x3:jug)]
  Delete Effects: [Holding(?x4:robot, ?x3:jug), JugAboveCup(?x3:jug, ?x1:cup), JugAboveCup(?x3:jug, ?x2:cup), RobotAboveCup(?x4:robot, ?x1:cup), RobotAboveCup(?x4:robot, ?x2:cup)]
  Ignore Effects: []
  Skill: PlaceJugInMachine(?x4:robot, ?x3:jug, ?x0:coffee_machine)
STRIPS-Op2:
  Parameters: [?x0:coffee_machine, ?x1:jug, ?x2:robot]
  Preconditions: [HandEmpty(?x2:robot), JugInMachine0(?x1:jug, ?x0:coffee_machine), JugPickable(?x1:jug), NotAboveCup(?x2:robot, ?x1:jug)]
  Add Effects: [JugFilled0(?x1:jug), PressingButton(?x2:robot, ?x0:coffee_machine)]
  Delete Effects: []
  Ignore Effects: []
  Skill: TurnMachineOnAndFill(?x2:robot, ?x0:coffee_machine)
STRIPS-Op3:
  Parameters: [?x0:coffee_machine, ?x1:cup, ?x2:cup, ?x3:jug, ?x4:robot]
  Preconditions: [HandEmpty(?x4:robot), JugFilled0(?x3:jug), JugInMachine0(?x3:jug, ?x0:coffee_machine), JugPickable(?x3:jug), NotAboveCup(?x4:robot, ?x3:jug), NotSameCup(?x1:cup, ?x2:cup), NotSameCup(?x2:cup, ?x1:cup), PressingButton(?x4:robot, ?x0:coffee_machine)]
  Add Effects: [Holding(?x4:robot, ?x3:jug), JugAboveCup(?x3:jug, ?x1:cup), JugAboveCup(?x3:jug, ?x2:cup), RobotAboveCup(?x4:robot, ?x1:cup), RobotAboveCup(?x4:robot, ?x2:cup)]
  Delete Effects: [HandEmpty(?x4:robot), JugInMachine0(?x3:jug, ?x0:coffee_machine), NotAboveCup(?x4:robot, ?x3:jug), PressingButton(?x4:robot, ?x0:coffee_machine)]
  Ignore Effects: []
  Skill: PickJug(?x4:robot, ?x3:jug)
STRIPS-Op4:
  Parameters: [?x0:cup, ?x1:jug, ?x2:robot]
  Preconditions: [Holding(?x2:robot, ?x1:jug), JugAboveCup(?x1:jug, ?x0:cup), JugFilled0(?x1:jug), JugPickable(?x1:jug), RobotAboveCup(?x2:robot, ?x0:cup)]
  Add Effects: [CupFilled(?x0:cup), CupReceivingLiquid0(?x0:cup)]
  Delete Effects: [JugFilled0(?x1:jug), JugPickable(?x1:jug)]
  Ignore Effects: []
  Skill: PourSomeLiquid(?x2:robot, ?x1:jug, ?x0:cup)
STRIPS-Op5:
  Parameters: [?x0:cup, ?x1:cup, ?x2:jug, ?x3:robot]
  Preconditions: [CupFilled(?x1:cup), CupReceivingLiquid0(?x1:cup), Holding(?x3:robot, ?x2:jug), JugAboveCup(?x2:jug, ?x0:cup), JugAboveCup(?x2:jug, ?x1:cup), NotSameCup(?x0:cup, ?x1:cup), NotSameCup(?x1:cup, ?x0:cup), RobotAboveCup(?x3:robot, ?x0:cup), RobotAboveCup(?x3:robot, ?x1:cup)]
  Add Effects: [CupFilled(?x0:cup), CupReceivingLiquid0(?x0:cup)]
  Delete Effects: [CupReceivingLiquid0(?x1:cup)]
  Ignore Effects: []
  Skill: PourSomeLiquid(?x3:robot, ?x2:jug, ?x0:cup)
STRIPS-Op6:
  Parameters: [?x0:jug, ?x1:robot]
  Preconditions: [HandEmpty(?x1:robot), NotAboveCup(?x1:robot, ?x0:jug), OnTable(?x0:jug)]
  Add Effects: [JugPickable(?x0:jug)]
  Delete Effects: []
  Ignore Effects: []
  Skill: RotateItemUntilHandleAccessible(?x1:robot, ?x0:jug)
STRIPS-Op7:
  Parameters: [?x0:cup, ?x1:jug, ?x2:robot]
  Preconditions: [HandEmpty(?x2:robot), JugPickable(?x1:jug), NotAboveCup(?x2:robot, ?x1:jug), OnTable(?x1:jug)]
  Add Effects: [Holding(?x2:robot, ?x1:jug), JugAboveCup(?x1:jug, ?x0:cup), RobotAboveCup(?x2:robot, ?x0:cup)]
  Delete Effects: [HandEmpty(?x2:robot), NotAboveCup(?x2:robot, ?x1:jug), OnTable(?x1:jug)]
  Ignore Effects: []
  Skill: PickJug(?x2:robot, ?x1:jug)
STRIPS-Op8:
  Parameters: [?x0:coffee_machine, ?x1:cup, ?x2:jug, ?x3:robot]
  Preconditions: [Holding(?x3:robot, ?x2:jug), JugAboveCup(?x2:jug, ?x1:cup), JugPickable(?x2:jug), RobotAboveCup(?x3:robot, ?x1:cup)]
  Add Effects: [HandEmpty(?x3:robot), JugInMachine0(?x2:jug, ?x0:coffee_machine), NotAboveCup(?x3:robot, ?x2:jug)]
  Delete Effects: [Holding(?x3:robot, ?x2:jug), JugAboveCup(?x2:jug, ?x1:cup), RobotAboveCup(?x3:robot, ?x1:cup)]
  Ignore Effects: []
  Skill: PlaceJugInMachine(?x3:robot, ?x2:jug, ?x0:coffee_machine)
STRIPS-Op9:
  Parameters: [?x0:coffee_machine, ?x1:cup, ?x2:jug, ?x3:robot]
  Preconditions: [HandEmpty(?x3:robot), JugFilled0(?x2:jug), JugInMachine0(?x2:jug, ?x0:coffee_machine), JugPickable(?x2:jug), NotAboveCup(?x3:robot, ?x2:jug), PressingButton(?x3:robot, ?x0:coffee_machine)]
  Add Effects: [Holding(?x3:robot, ?x2:jug), JugAboveCup(?x2:jug, ?x1:cup), RobotAboveCup(?x3:robot, ?x1:cup)]
  Delete Effects: [HandEmpty(?x3:robot), JugInMachine0(?x2:jug, ?x0:coffee_machine), NotAboveCup(?x3:robot, ?x2:jug), PressingButton(?x3:robot, ?x0:coffee_machine)]
  Ignore Effects: []
  Skill: PickJug(?x3:robot, ?x2:jug)
STRIPS-Op10:
  Parameters: [?x0:cup, ?x1:jug, ?x2:robot]
  Preconditions: [Holding(?x2:robot, ?x1:jug), JugAboveCup(?x1:jug, ?x0:cup), JugFilled0(?x1:jug), JugPickable(?x1:jug), RobotAboveCup(?x2:robot, ?x0:cup)]
  Add Effects: [CupFilled(?x0:cup)]
  Delete Effects: [JugFilled0(?x1:jug), JugPickable(?x1:jug)]
  Ignore Effects: []
  Skill: PourSomeLiquid(?x2:robot, ?x1:jug, ?x0:cup)
\end{lstlisting}
\end{tcolorbox}

\begin{tcolorbox}[
    title={\textbf{Example learned predicates and learned action models in Juice}},
    colback=gray!10,
    colframe=black,
    sharp corners=southwest,
    boxrule=0.8pt,
    breakable
  ]
  \lstset{
    basicstyle=\ttfamily\scriptsize,
    breaklines=true
  }
  \begin{lstlisting}
Learned predicates:
IsCup0
NearJuiceValve0
InsideWasteValveRegion0
HoldingContainer0
OutsideJuiceMachine0
LidClosed0
HoldingObj0
Functional0
JuiceMachineOpen0
IsPlastic0
HandEmpty0
HasContents0

Learned operators:
STRIPS-Op0:
  Parameters: [?x0:robot, ?x1:juicer]
  Preconditions: [Functional0(?x1:juicer), HandEmpty0(?x0:robot), JuiceMachineOpen0(?x1:juicer)]
  Add Effects: [LidClosed0(?x1:juicer)]
  Delete Effects: [JuiceMachineOpen0(?x1:juicer)]
  Ignore Effects: []
  Skill: CloseLid(?x0:robot, ?x1:juicer)

STRIPS-Op1:
  Parameters: [?x0:robot, ?x1:container]
  Preconditions: [HandEmpty0(?x0:robot), IsCup0(?x1:container)]
  Add Effects: [HoldingContainer0(?x0:robot, ?x1:container)]
  Delete Effects: [HandEmpty0(?x0:robot)]
  Ignore Effects: []
  Skill: PickContainer(?x0:robot, ?x1:container)

STRIPS-Op2:
  Parameters: [?x0:robot, ?x1:juicer, ?x2:container]
  Preconditions: [Empty(?x2:container), Functional0(?x1:juicer), HoldingContainer0(?x0:robot, ?x2:container), IsCup0(?x2:container), LidClosed0(?x1:juicer)]
  Add Effects: [HandEmpty0(?x0:robot), NearJuiceValve0(?x2:container, ?x1:juicer)]
  Delete Effects: [HoldingContainer0(?x0:robot, ?x2:container)]
  Ignore Effects: []
  Skill: PlaceInsideJuiceValveRegion(?x0:robot, ?x2:container, ?x1:juicer)

STRIPS-Op5:
  Parameters: [?x0:robot, ?x1:juicer, ?x2:container]
  Preconditions: [Empty(?x2:container), Functional0(?x1:juicer), HoldingContainer0(?x0:robot, ?x2:container), IsCup0(?x2:container), IsPlastic0(?x2:container)]
  Add Effects: [HandEmpty0(?x0:robot), InsideWasteValveRegion0(?x2:container, ?x1:juicer)]
  Delete Effects: [HoldingContainer0(?x0:robot, ?x2:container)]
  Ignore Effects: []
  Skill: PlaceInsideWasteValveRegion(?x0:robot, ?x2:container, ?x1:juicer)

STRIPS-Op7:
  Parameters: [?x0:container, ?x1:container, ?x2:robot]
  Preconditions: [Empty(?x0:container), HasContents0(?x1:container), HoldingContainer0(?x2:robot, ?x1:container), IsCup0(?x1:container), IsPlastic0(?x1:container)]
  Add Effects: [Empty(?x1:container), HasContents0(?x0:container)]
  Delete Effects: [Empty(?x0:container), HasContents0(?x1:container)]
  Ignore Effects: []
  Skill: DumpFromOneIntoOther(?x2:robot, ?x1:container, ?x0:container)

STRIPS-Op10:
  Parameters: [?x0:container, ?x1:container, ?x2:robot, ?x3:juicer, ?x4:movable_object]
  Preconditions: [Empty(?x0:container), Empty(?x1:container), Functional0(?x3:juicer), HandEmpty0(?x2:robot), Inside(?x4:movable_object, ?x3:juicer), InsideWasteValveRegion0(?x1:container, ?x3:juicer), IsCup0(?x0:container), IsCup0(?x1:container), IsPlastic0(?x1:container), LidClosed0(?x3:juicer), NearJuiceValve0(?x0:container, ?x3:juicer)]
  Add Effects: [HasContents0(?x0:container), HasContents0(?x1:container), JuiceInCup(?x4:movable_object, ?x0:container), OutsideJuiceMachine0(?x4:movable_object)]
  Delete Effects: [Empty(?x0:container), Empty(?x1:container), Inside(?x4:movable_object, ?x3:juicer)]
  Ignore Effects: []
  Skill: TurnOnAndRunMachine(?x2:robot, ?x3:juicer, ?x0:container, ?x1:container)

STRIPS-Op11:
  Parameters: [?x0:robot, ?x1:movable_object]
  Preconditions: [HandEmpty0(?x0:robot), OutsideJuiceMachine0(?x1:movable_object)]
  Add Effects: [HoldingObj0(?x0:robot, ?x1:movable_object)]
  Delete Effects: [HandEmpty0(?x0:robot)]
  Ignore Effects: []
  Skill: PickMovable(?x0:robot, ?x1:movable_object)

STRIPS-Op12:
  Parameters: [?x0:robot, ?x1:juicer, ?x2:movable_object]
  Preconditions: [HoldingObj0(?x0:robot, ?x2:movable_object), JuiceMachineOpen0(?x1:juicer), OutsideJuiceMachine0(?x2:movable_object)]
  Add Effects: [HandEmpty0(?x0:robot), Inside(?x2:movable_object, ?x1:juicer)]
  Delete Effects: [HoldingObj0(?x0:robot, ?x2:movable_object), OutsideJuiceMachine0(?x2:movable_object)]
  Ignore Effects: []
  Skill: PlaceInside(?x0:robot, ?x2:movable_object, ?x1:juicer)
\end{lstlisting}
\end{tcolorbox}

\begin{tcolorbox}[
    title={\textbf{Example learned predicates and learned action models in Cleanup}},
    colback=gray!10,
    colframe=black,
    sharp corners=southwest,
    boxrule=0.8pt,
    breakable
  ]
  \lstset{
    basicstyle=\ttfamily\scriptsize,
    breaklines=true
  }
  \begin{lstlisting}
Learned predicates:
IsOnFloor0
IsEraser0
Holding0
NoObjectsOnTop1
IsOn0
HandEmpty0

Learned operators:
STRIPS-Op0:
  Parameters: [?x0:movable_object, ?x1:table, ?x2:robot]
  Preconditions: [HandEmpty0(?x2:robot), IsOn0(?x0:movable_object, ?x1:table)]
  Add Effects: [Holding0(?x2:robot, ?x0:movable_object), NoObjectsOnTop1(?x1:table)]
  Delete Effects: [HandEmpty0(?x2:robot), IsOn0(?x0:movable_object, ?x1:table)]
  Ignore Effects: []
  Skill: PickFromTop(?x2:robot, ?x0:movable_object, ?x1:table)

STRIPS-Op1:
  Parameters: [?x0:movable_object, ?x1:robot, ?x2:trash_can]
  Preconditions: [Holding0(?x1:robot, ?x0:movable_object)]
  Add Effects: [HandEmpty0(?x1:robot), InsideContainer(?x0:movable_object, ?x2:trash_can)]
  Delete Effects: [Holding0(?x1:robot, ?x0:movable_object)]
  Ignore Effects: []
  Skill: PlaceInside(?x1:robot, ?x0:movable_object, ?x2:trash_can)

STRIPS-Op2:
  Parameters: [?x0:movable_object, ?x1:robot]
  Preconditions: [HandEmpty0(?x1:robot), IsOnFloor0(?x0:movable_object)]
  Add Effects: [Holding0(?x1:robot, ?x0:movable_object)]
  Delete Effects: [HandEmpty0(?x1:robot), IsOnFloor0(?x0:movable_object)]
  Ignore Effects: []
  Skill: PickFromFloor(?x1:robot, ?x0:movable_object)

STRIPS-Op3:
  Parameters: [?x0:table, ?x1:movable_object, ?x2:robot]
  Preconditions: [Holding0(?x2:robot, ?x1:movable_object), IsEraser0(?x1:movable_object), NoObjectsOnTop1(?x0:table)]
  Add Effects: [WipedOfMarkerScribbles(?x0:table)]
  Delete Effects: []
  Ignore Effects: []
  Skill: WipeAndContinueHoldingEraser(?x2:robot, ?x1:movable_object, ?x0:table)

STRIPS-Op7:
  Parameters: [?x0:movable_object, ?x1:robot, ?x2:trash_can]
  Preconditions: [HandEmpty0(?x1:robot), InsideContainer(?x0:movable_object, ?x2:trash_can)]
  Add Effects: [IsOnFloor0(?x0:movable_object)]
  Delete Effects: [InsideContainer(?x0:movable_object, ?x2:trash_can)]
  Ignore Effects: []
  Skill: DumpContentsOntoFloor(?x1:robot, ?x2:trash_can)
\end{lstlisting}
\end{tcolorbox}




\end{document}